\documentclass{article}

\usepackage{PRIMEarxiv}

\usepackage{times}
\usepackage{soul}
\usepackage[utf8]{inputenc} 
\usepackage[T1]{fontenc}    
\usepackage[hidelinks]{hyperref}       
\usepackage{url}            
\usepackage{booktabs}       
\usepackage{amsfonts}       
\usepackage{nicefrac}       
\usepackage{microtype}      
\usepackage{lipsum}
\usepackage{fancyhdr}       
\usepackage{graphicx}       
\usepackage[font = small]{caption}
\usepackage{graphicx}
\usepackage{amsmath}
\usepackage{amsthm}
\usepackage{thmtools} 
\usepackage{thm-restate}
\usepackage{amssymb}
\usepackage{booktabs}
\usepackage{algorithm}
\usepackage{xspace}
\usepackage[noend]{algpseudocode}
\usepackage{proof}
\usepackage{tikz}
\usepackage{wrapfig}
\usepackage{todonotes}
\usepackage{cleveref}
\usepackage{subcaption}
\usepackage{marvosym}
\usetikzlibrary{positioning}
\usetikzlibrary{decorations.pathmorphing}

\usepackage{fontawesome5}
\graphicspath{{media/}}     

\newcommand\ldiamondarg[1]{\langle#1\rangle}
\newcommand\ldiaarg[1]{\langle#1\rangle}

\newcommand{\LL}{\mathcal{L}}
\newcommand{\M}{\mathcal{M}}

\newcommand{\cP}{\mathcal{P}}

\newcommand{\cS}{\mathcal{S}}

\newcommand{\B}{\ensuremath{\mathbb{B}}}
\newcommand{\Bb}{\ensuremath{\mathcal{B}}}

\newcommand{\Ag}{Ag}
\newcommand{\BP}{\mathcal{P}}

\renewcommand{\phi}{\varphi}

\newcommand{\Exp}{Exp} 

\newcommand{\POL}{\mathsf{POL}}
\newcommand{\DPDL}{\mathsf{DPDL}}

\newcommand{\DEL}{\mathsf{DEL}}
\newcommand{\BAPAL}{\mathsf{BAPAL}}
\newcommand{\PDL}{\mbox{\rm PDL}}
\newcommand{\EPL}{\mbox{\rm EPL}}

\newcommand{\PAL}{\mathsf{PAL}}

\newcommand{\union}{\cup}

\newcommand\setprop{\mathcal{P}}

\newcommand{\MD}{M}

\newcommand{\modelM}{\mathcal M}

\newcommand{\set}[1]{\{#1\}}

\newcommand{\EXPTIME}{\mathsf{EXPTIME}}
\newcommand{\AEXPSPACE}{\mathsf{AEXPSPACE}}

\newcommand{\regdiv}[1]{\ensuremath{\backslash} #1}
\newcommand{\LTLK}{\mathsf{LTL_K}}

\newtheorem{theorem}{Theorem}
\newtheorem{example}[theorem]{Example}
\newtheorem{proposition}[theorem]{Proposition}
\newtheorem{corollary}[theorem]{Corollary}

\newtheorem{claim}[theorem]{Claim}
\newtheorem{definition}[theorem]{Definition}

\newtheorem{observation}[theorem]{Observation}

\newcommand{\anyproblem}{A}
\newcommand{\tm}{M}
\newcommand{\limply}{\rightarrow}
\newcommand{\lequiv}{\leftrightarrow}
\newcommand{\lbox}{\square}
\newcommand{\ldia}{\Diamond}
\newcommand{\bigxor}{\bigoplus}
\newcommand{\setsymbols}{Sym}
\newcommand{\qacc}{q_{acc}}
\newcommand{\qrej}{q_{rej}}

\newcommand{\obssymbolfirstconfig}[1]{1{:}#1}
\newcommand{\obssymbolsecondconfig}[1]{2{:}#1}
\newcommand{\obssymbolthirdconfig}[1]{3{:}#1}
\newcommand{\formulasymbolfirstconfig}[1]{\ldiaarg{1{:}#1}\top}
\newcommand{\formulasymbolsecondconfig}[1]{\ldiaarg{2{:}#1}\top}
\newcommand{\formulasymbolthirdconfig}[1]{\ldiaarg{3{:}#1}\top}
\newcommand{\formulasymboliconfig}[1]{\ldiaarg{i{:}#1}\top}
\newcommand{\formulasymboljconfig}[1]{\ldiaarg{j{:}#1}\top}
\newcommand{\successorfunctiona}{\mathsf{succ}_a}
\newcommand{\successorfunctionb}{\mathsf{succ}_b}
\newcommand{\leftmostrightmostsymbol}{\#}

\newcommand{\transitionprogram}{a {\cup} b}
\newcommand{\transitionsprogram}{(a {\cup} b)^*}
\newcommand{\positionfirstconfig}{pos_1}
\newcommand{\positioniconfig}{pos_i}
\newcommand{\positionjconfig}{pos_j}
\newcommand{\positionsecondconfig}{pos_2}
\newcommand{\positionthirdconfig}{pos_3}
\newcommand{\symbolfirstconfig}{\alpha}
\newcommand{\symbolsecondconfig}{\beta}
\newcommand{\symbolthirdconfig}{\gamma}

\newcommand{\propwin}{\ldiaarg{win}\top}
\newcommand{\formulaterminal}{\mathsf{final}}
\newcommand{\formulaaccepted}{\mathsf{accepted}}
\newcommand{\formularejected}{\mathsf{rejected}}

\newcommand{\configurationdraw}[3]{
\footnotesize
\begin{tikzpicture}[xscale=0.7]
\node[cell] at (0, 0) {#1};
\node[cell] at (1, 0) {#2};
\node[cell] at (2, 0) {#3};
\node[cell] at (3, 0) {};
\end{tikzpicture}
}

\newcommand{\leaf}{t}

\tikzstyle{bubble} = [fill=gray!10!white,decorate, decoration={snake, segment length=2mm, amplitude=0.1mm}, line width=0.3mm, draw, inner sep=1mm]
\tikzstyle{obstransition} = [line width=0.3mm, -latex]
\tikzstyle{hintikkastate} = [draw, inner sep=0.5mm]
\usetikzlibrary{patterns,snakes}

\title{Reasoning about knowledge on regular expressions is 2EXPTIME-complete 
}

\author{
  Avijeet Ghosh \\
  Chennai Mathematical Institute \\
  Chennai, India\\
  \texttt{avi.ghosh23@gmail.com} \\
   \And
  Sujata Ghosh \\
  Indian Statistical Institute \\
  Chennai, India \\
  \texttt{sujata@isichennai.res.in} \\
    \And 
  Fran{\c{c}}ois Schwarzentruber \\
  ENS De Lyon \\ 
  France \\
  \texttt{francois.schwarzentruber@ens-lyon.fr}
}



\algnewcommand\algorithmicforeach{\textbf{for each}}
\algdef{S}[FOR]{ForEach}[1]{\algorithmicforeach\ #1\ \algorithmicdo}

\begin{document}
\maketitle

\begin{abstract}
Logics for reasoning about knowledge and actions have seen many applications in various domains of multi-agent systems, including epistemic planning. Change of knowledge based on observations about the surroundings forms a key aspect in such planning scenarios. Public Observation Logic ($\POL$) is a variant of public announcement logic for reasoning about knowledge that gets updated based on public observations. Each state in an epistemic (Kripke) model is equipped with a set of expected observations. These states evolve as the expectations get matched with the actual observations. In this work, 
we prove that the satisfiability problem of $\POL$ is 2EXPTIME-complete.
\end{abstract}

\keywords{Public observation logic \and
Propositional Dynamic Logic \and
Satisfiability problem \and
Complexity}

\section{Introduction}
\label{section:introduction}

\newcommand{\Avijeet}[1]{\textcolor{blue}{#1}}

Intelligent artificial agents are being used for performing various tasks including planning and scheduling in real life, from simple to more complicated ones. For example, a robot may try to move from a point to another by overcoming certain hurdles. Or, it may try to keep an eye over the surroundings without other agents \textit{knowing} about it. Accomplishing a goal in such surveillance activities may involve the robot's knowledge about other agents' knowledge. Automated planning~\cite{ghallabplanning2004} is a branch of study in multi-agent systems that involves deciding whether a sequence or plan of actions exists to attain some goal. An extension of such planning studies is termed as \textit{epistemic planning}~\cite{DBLP:journals/corr/Bolander17}, 
where the goal, like in the case of surveillance robot, involves knowledge of multiple agents.

Reasoning about knowledge using logical systems has been studied extensively in the domain of modal logic \cite{modallogicblackburn}, more specifically using epistemic logic \cite{faginknowledge95}. In addition, reasoning about the knowledge dynamics of agents has been studied using dynamic epistemic logic ($\DEL$)~\cite{DEL}, among others. Evidently, a popular approach in epistemic planning is the use of model-checking problem in $\DEL$ \cite{DBLP:journals/corr/Bolander17}. However, this problem turns out to be undecidable when one considers finite iterations of such actions in $\DEL$~\cite{aucher-bolander13}. 
Moreover, it is shown that for the general epistemic planning tasks, the plan existence problem is already undecidable with two agents.

Public observation logic ($\POL$)~\cite{vanhidden2014} deals with \textit{expected} observations (actions) 
that are associated with each state in an epistemic model \cite{faginknowledge95}, and are represented by 
regular expressions. The model gets updated depending on the matching of actual and expected observations, and accordingly, agents' knowledge gets updated as well. 
This dynamic behavior based on some sequence (finite iteration) of observations inherently makes this setting useful in reasoning about various concepts involving knowledge and actions, for example, epistemic planning. 

In \cite{krpolsf23}, we show that public announcement logic ($\PAL$) with propositional announcements is closely related to the {\em word} fragment of $\POL,$ where the regular expressions describing observations are only in the word form. In this sense, $\POL$ can be considered as a dynamic logic-like extension of $\PAL$, taking iteration of the Boolean announcements under its wings. Thus, it is worthwhile to check whether $\POL$ is decidable, since $\PAL$ with iterated announcements is undecidable~\cite{DBLP:journals/sLogica/MillerM05}.  Note that the decidability~\cite{DBLP:journals/lmcs/DitmarschF22} of arbitrary $\PAL$ with Boolean announcements ($\BAPAL$) provides a push towards an affirmative response to our query. Before moving forward, let us provide a scenario that can be modeled by $\POL.$

\begin{example}
Consider a surveillance drone hovering over the boundary zone between two territories, $T_1$ and $T_2$, say, in conflict with each other. Suppose the drone is deployed by $T_1$, and if it is detected in the airspace of $T_2$, it might get destroyed. How would the drone differentiate between the two territories so that it can restrict itself from entering $T_2$? According to its expectations based on the vegetation in the area, if it observes the sequence of (spruce$^\ast$-pine$^\ast$-cedar-fir$^\ast$)$^\ast$ ($\ast$ denotes the continuance of such sequences), it would know that it is in area $T_1$, while if it observes (spruce$^\ast$-pine$^\ast$-larch-fir$^\ast$)$^\ast,$ it would know that it is in $T_2$.
\end{example}

With regard to the computational behavior of $\POL,$ we explored the complexity of the model-checking problem for $\POL$ in \cite{polijcai22modelcheck}. In addition, we investigated the satisfiability problem for the star-free fragment of $\POL$ in \cite{krpolsf23}, where the observations comprise of star-free regular expressions. In this work, we complete this study by providing an answer to the remaining open problem concerning the decidability of full $\POL.$ 
We show that the satisfiability problem of $\POL$ (with Kleene star) is 2EXPTIME-complete, in contrast to $\PAL$ with iterated announcements, making it a more viable option for modelling planning and related problems. 
The techniques used in our proofs may be of a more general interest.

For the upper bound, we start with the usual filtration argument \cite{modallogicblackburn} - if a formula is satisfiable then it is satisfiable in a model of exponential size.
 Although filtration is usually sufficient to prove decidability, for $\POL$ this is not the case, as the expected observations associated with each state in a model may be arbitrary (Section~\ref{sec:finitemodelproperty}). We characterize satisfiability in terms of a finite syntactic structure defined by Hintikka sets \cite{smullyan1995first},
which we call a \emph{bubble transition structure}.
  We provide a correspondence between the expected observations and the corresponding automata (Section~\ref{sec:decidability}) to facilitate our study. 
  Then in exponential time, we reduce the satisfiability problem of deterministic propositional dynamic logic ($\DPDL$) to that of $\POL$ by encoding the constraints on the epistemic structures by propositional theories (Section~\ref{sec:reductiontodpdl}).
For the lower bound, we encode a superposition of \emph{three} same configurations of a Turing machine in the leaves of a $\POL$ model considered as a full binary tree (Section~\ref{sec:lowerbound}). Without further ado, let us start by recalling $\POL$.

\section{Public Observation Logic}
\label{section:background}
To begin with, we provide a brief overview of Public Observation Logic \cite{vanhidden2014}. Let $\BP$ be a countable set of propositional letters, 
$\Ag$ be a finite set of agents, and $\Sigma$ be a finite alphabet of atomic actions/observations. We now introduce \emph{observation expressions} as follows:

\begin{definition}[Observation Expression]
    Given a finite set of observations $\Sigma$, \emph{observation expressions} are defined recursively as:
  $  \pi := \emptyset\ \  |\ \  a\ \  |\ \  \pi + \pi\ \ |\ \ \pi;\pi\ \ |\ \ \pi^\star,$
with $a \in \Sigma$. 
\end{definition}

Note that an observation expression $\pi$ is 
a regular expression, and 
$\LL(\pi)$ denotes the language corresponding to the regular expression $\pi$.
We use these observation expressions to describe observations within formulas (cf. Definition~\ref{defi:polsyntax}) as well as expected observations at states in a model (cf. Definition~\ref{defi:polmodel}). 
In the surveillance example, an observation may be \emph{spruce-pine-cedar-fir-spruce} (abbreviated as $\mathit{spcfs}$) or \emph{spruce-pine-larch-fir-spruce} (abbreviated as $\mathit{splfs}$), among others. Let us now describe the language of $\POL.$

\begin{definition} [$\POL$ Syntax]\label{defi:polsyntax}
The language of $\POL$ can be recursively defined as:
\begin{align*}
    \varphi := \top\ \  |\ \  p\ \  |\ \  \varphi \vee \varphi\ \ |\ \ \neg{\varphi}\ \ |\ \ \hat{K_i}\varphi\ \ |\ \ \ldiamondarg\pi\varphi
\end{align*}
where $p \in \BP$, $i \in \Ag$, 
and $\pi$ is an observation expression over $\Sigma$.
\end{definition}
The box formulas are defined as follows:
$[\pi]\varphi = \neg{\ldiamondarg\pi\neg{\varphi}}$, $
    K_i \varphi = \neg{\hat{K_i}\neg{\varphi}}$. 
The formula $K_i \varphi $ is read as `agent $i$ knows $\varphi$', while 
 $\hat{K_i}\varphi$ is read as `agent $i$ considers $\varphi$ as an epistemic possibility'. The formula $\ldiamondarg\pi\psi$ expresses that there is a sequence of atomic observations that matches the language of the (regular) expression $\pi$ and $\psi$ holds after the said sequence is observed publicly.
For example, the formula $\ldiamondarg{(s^\star p^\star l f^\star) c}K_dT_2$ expresses that after an observation of finite sequences of spruce and pine followed by a larch and then a finite sequence of fir, the drone knows that it is in the region $T_2.$ We are now ready to describe the {\em $\POL$ models} 
	\cite{vanhidden2014} that capture the expected observations of agents. 
	They can be seen as epistemic models \cite{faginknowledge95} together with, for each world, a set of potential observations.

\begin{definition}[$\POL$ model]\label{defi:polmodel}
A $\POL$ model is a tuple $\mathcal{M} = (S, R, V, \Exp)$, where, (i) $S$ is a non-empty set of states, (ii) $R_i \subseteq S\times S$ is an equivalence relation for all $i\in \Ag$. $R = \{R_i\}_{i\in\Ag}$, (iii)  $V\colon S\rightarrow 2^{\BP}$ is a valuation function, and (iv) $\Exp\colon S\rightarrow RE_\Sigma$ is an expectation function assigning an observation expression over $\Sigma$ to each state in $S$.

\end{definition}

\tikzstyle{world} = [draw]
\begin{figure} 
		\begin{center}
			\begin{tikzpicture}
				\node[world] (u) {$T_1, \lnot T_2$};
				\node[world] (v) at (4, 0) {$\lnot T_1, T_2$};
				\node[left = 0mm of u] {$u$};
				\node[right = 0mm of v] {$v$};
				\node[below = 0mm of u] {$(s^*p^*cf^*)^*$};
				\node[below = 0mm of v] {$(s^*p^*lf^*)^*$};
				\draw (u) edge node[above] {$Drone$} (v);
			\end{tikzpicture}
		\end{center}
		\vspace{-.5cm}
		\caption{$\M_{sd}$ (the surveillance drone model)}\label{figure:drone}
	\end{figure}
    
Figure 1 models the scenario discussed in the introduction. The model consists of the set of states $S = \{u, v \}$, with $u$ representing the state of the drone hovering over the region $T_1$ and $v$ representing the same for $T_2$. The expected observations of the drone in the two states are assigned accordingly (cf. the regular expressions assigned to $u$ and $v$).




To interpret the changes in agent knowledge based on observations, $\POL$ semantics involves model updates. The observation expressions associated with the states get updated according to the sequence of actions already observed. To model this idea formally, we first explain the process of \textit{residuation} of observation (regular) expressions with respect to words: 
Given a word $w\in\Sigma^\star$ and a regular expression $\pi$ over $\Sigma$, $\pi\regdiv w$ is a regular expression, called the \textit{residuation} of $\pi$ with $w$, where, $\LL(\pi\regdiv w) = \{u\mid wu\in\LL(\pi)\}$. For example, for $a, b \in \Sigma,$ $(b^\star aa(a + b)^\star)\regdiv ba = a(a+b)^\star$. 

\begin{definition}[Model update by observation]
Given a word $w$ over $\Sigma$, the model $\mathcal{M}|_w = (S', R', V', Exp')$ is defined as follows: (i) $S' = \{ s\in S\ \ |\ \  \mathcal{L}(Exp(s)\backslash w)\neq \emptyset\}$, (ii) $R' = R|_{S'\times S'}$, (iii) $V' = V|_{S'\times S'}$, and, (iv) $Exp'$ is given by $Exp'(s) = Exp(s)\backslash w$ for all $s\in S'$.
\end{definition}

    
    
    

We are now ready to give the interpretation of the $\POL$ formulas in $\POL$ models.

\begin{definition}[Truth of a POL formula]
\label{definition:POLtruthconditions}
Given a model $\mathcal{M} = (S, R, V, Exp)$ and an $s\in S$, the truth definition of a POL formula $\varphi$ ($\mathcal{M}, s\vDash \varphi$), is 
given as follows:
\begin{itemize}
    \item $\mathcal{M},s\vDash p$ iff $p\in V(s)$, where $p\in \BP$.

    \item $\mathcal{M},s\vDash \neg{\psi}$ iff $\mathcal{M}, s\nvDash \psi$.

 \item $\mathcal{M},s\vDash \psi\vee \chi$ iff $\mathcal{M},s\vDash \psi$ or $\mathcal{M},s\vDash \chi$.
    
    \item $\mathcal{M},s \vDash \hat{K_i}\psi$ iff there is $t\in S$, s. th. $s R_i t$ and $\mathcal{M},t\vDash \psi$.
    
    \item $\mathcal{M},s \vDash \ldiamondarg{\pi}\psi$ iff there exists $w\in \mathcal{L}(\pi)$ such that $\mathcal{L}(Exp(s)\backslash w)\neq \emptyset$ and $\mathcal{M}|_w ,s\vDash \psi$.
\end{itemize}
\end{definition}

The truth definitions are as usual except for the last one. The formula $\ldiamondarg{\pi}\psi$ holds if there is an observation sequence (a word $w$, say) that matches $\pi$, and after observing $w$ (publicly), $\psi$ holds. The dual formula $[\pi]\psi$ is interpreted accordingly. As for the drone example and its model (Figure~\ref{figure:drone}) defined above, we can verify: 

\begin{itemize}
    \item[-] $\M_{sd}, s \models [s^*p^*]\neg(K_d T_1 \lor K_d \neg T_1)$. This example corresponds to the drone being uncertain about its whereabouts: observing an arbitrary number of $s$'s followed by $p$'s is compatible with both the expectation $(s^*p^*cf^*)^*$ of the $T_1$ vegetation, and the expectation $(s^*p^*lf^*)^*$ of the non-$T_1$ vegetation.
			
	\item[-] $\M_{sd}, s \models \ldiamondarg{s^*p^*c}(K_d T_1)$. This example expresses the existence of a sequence of observations that reveals that the drone is in the region $T_1$.
\end{itemize}

\noindent\textbf{Satisfiability Problem.} The satisfiability problem of $\POL$ is as follows: Given a $\POL$ formula $\varphi$, does there exist a $\POL$ model $\M$ and a state $s$ in it such that $\M,s\vDash\phi$? 
In what follows, we show that the $\POL$ satisfiability problem is decidable and explore the complexity of the problem.



\section{Finite model property}
\label{sec:finitemodelproperty}
As a first step
we show the following result:

\begin{theorem}
    [Finite model property]
    If $\varphi$ is satisfiable then $\varphi$ is satisfied in a $\POL$ model with $2^{O(|\varphi|)}$ states.
\end{theorem}

\subsection{Closure Sets}
\newcommand{\FL}{F\!L}

\begin{definition}[Fischer-Ladner Closure]
    \label{defi:FLclosure}
     The Fisher-Ladner closure of a formula $\varphi$, denoted by $\FL(\varphi)$, is the smallest set containing $\varphi$ and satisfying the following conditions:
    \begin{itemize}
        
        \item if $\psi\in \FL(\varphi)$ and $\psi$ not starting with $\neg$ then $\neg\psi \in \FL(\varphi)$ 
        
        \item if $\neg\psi, K_i\psi, \hat{K_i}\psi, [\pi]\psi$ or $\ldiamondarg{\pi}\psi$ in $\FL(\varphi)$ then 
        $\psi {\in}\FL(\varphi)$ 

        \item if $\psi\wedge\chi$ or $ \psi\vee\chi$ are in $FL(\varphi)$ then $\psi,\chi\in FL(\varphi)$ 
        
        \item if $\ldiamondarg{\pi_1;\pi_2}\psi \in \FL(\varphi)$ then $\ldiamondarg{\pi_1}\ldiamondarg{\pi_2}\psi \in FL(\varphi)$
        
        \item if $\ldiamondarg{\pi_1+\pi_2}\psi \in \FL(\varphi)$ then  $\ldiamondarg{\pi_1}\psi, \ldiamondarg{\pi_2}\psi \in \FL(\varphi)$
        
        \item if $\ldiamondarg{\pi^{\star}}\psi \in FL(\varphi)$ then $\ldiamondarg{\pi}\ldiamondarg{\pi^\star}\psi \in FL(\varphi)$
        
        \item if $[\pi_1;\pi_2]\psi \in FL(\varphi)$ then $[\pi_1][\pi_2]\psi \in FL(\varphi)$
        
        \item if $[\pi_1+\pi_2]\psi \in FL(\varphi)$ then $[\pi_1]\psi, [\pi_2]\psi\in FL(\varphi)$
        
        \item if $[\pi^{\star}]\psi \in FL(\varphi)$ then $[\pi][\pi^\star]\psi \in FL(\varphi)$ 
    \end{itemize}
\end{definition}
 In simple words, the Fischer-Ladner closure of a formula constitutes all subformulas that need to be considered in a satisfiability argument to satisfy the original formula.  Note that if $[\pi]\psi\in FL(\varphi)$, we have $[\pi\regdiv a]\psi\in FL(\varphi)$. For example, suppose $[b^\star ab]\psi\in FL(\varphi)$. Note that $b^\star ab\regdiv a = b$. By definition~\ref{defi:FLclosure}, $[b^\star][ab]\psi\in FL(\varphi)$ implies $[ab]\psi, [b][b^\star][a]\psi\in FL(\varphi)$. Since $[ab]\psi\in FL(\varphi)$, therefore $[a][b]\psi\in FL(\varphi)$, which finally implies $[b]\psi\in FL(\varphi)$. The same is also true for the diamond formulas ($\ldiamondarg{\pi}\psi$) as well.
\begin{example}
  Consider $\varphi = \ldiamondarg{(a+b)^\star}p$.
  Then, we have: $
     FL(\varphi) = \{\varphi, \neg \varphi, p, \ldiamondarg{a+b}\ldiamondarg{(a+b)^\star}p, \ldots \} $.    
\end{example}

\begin{observation}{\cite{hareldynamiclogic}}\label{obs:linearflclosure}
    Given $\varphi$, $|FL(\varphi)|\leq O(|\varphi|)$.
\end{observation}


\subsection{Filtration}

A standard 
approach for showing decidability of the satisfiability problem in modal logics is the filtration technique \cite{modallogicblackburn}, which goes by proving the small model property: For any satisfiable formula, there exists a finite model satisfying it whose size can be bounded with respect to the input formula. We provide a similar argument using the following construction of the \textit{small model}. We first define an equivalence relation $\sim$ among the states of a model $\M = \ldiamondarg{S, \{R_i\}_{i\in \Ag}, V, Exp}$ ($\sim\subseteq S\times S$) with respect to a formula $\varphi$ as follows:
\begin{align*}
        s \sim s' &\mbox{ iff for all } \psi \in FL(\varphi), (\M, s \models \psi \mbox{ iff } \M, s' \models \psi)
\end{align*}
Note that, the relation $\sim$ is reflexive, transitive and symmetric, that is, an equivalence relation over $S$. For any $s\in S$, we denote $[s]$  to be the equivalence class with respect to $\sim$ that contains $s$. Next we give the small model construction following~\cite{modallogicblackburn}. 

\begin{definition}[Small Model of a Formula]
\label{def:filtratedmodel}
    Given a model $\M = \ldiamondarg{S, \{R_i\}_{i\in \Ag}, V, Exp}$ and a formula $\varphi$, a small model $\M^\sim = \ldiamondarg{S^\sim,
    \{R^\sim_i\}_{i\in \Ag}, V^\sim, Exp^\sim}$ is defined as:
    \begin{itemize}
        \item $S^\sim = \{[s]\mid s\in S\}$
        \item $([s], [s'])\in R^\sim_i$ if these conditions hold:
        \begin{enumerate}
            \item there exists $s_1\in [s]$ and $s_2\in [s']$ s. t. $(s_1, s_2)\in R_i$.
            \item for all $\hat{K}_i\psi\in FL(\varphi)$, 
            $\M,s'\vDash\psi\vee\hat{K}_i\psi$ implies that $\M,s\vDash\hat{K}_i\psi$.
        \end{enumerate}
        \item $V^\sim([s]) = V(s)$
        \item $Exp^\sim([s]) = Exp(s^c)$ for some $s^c\in [s]$.
    \end{itemize}
\end{definition}

Note that the above relation $R^\sim_i\subseteq S^\sim\times S^\sim$ is an equivalence relation. The trickier parts are the transitivity and symmetry. The former can be proved using condition (2), as discussed in \cite{modallogicblackburn}. The latter, that is symmetry, can be proved using (1) and (2). 
 Before proving the small model property, we prove the following lemma, which shows that the choice of $s^c$ in Definition~\ref{def:filtratedmodel} is not important. 
 The main theorem follows.
 \begin{restatable}{lemma}{expfinitemodel}
     For any formula of the form $\psi' = \ldiamondarg{\pi}\psi\in FL(\varphi)$, if $s\sim s'$, then there exists a $w\in\LL(\pi)$ such that [($\M,s\vDash\psi'$, $s$ survives in $\M|_w$ and $\M|_w,s\vDash\psi$) iff ($\M,s'\vDash\psi'$, $s'$ survives in $\M|_w$ and $\M|_w,s'\vDash\psi$)].
 \end{restatable}
 \noindent\textbf{Proof Sketch.} The proof goes by induction on the size of $\pi$. Since $s$ and $s'$ satisfy the same formulas in $FL(\varphi)$, the result follows by definition~\ref{defi:FLclosure}. In particular, the inclusion of formulas within $\FL(\varphi)$ that take care of the prefixes of $\pi$ in $\ldiamondarg{\pi}\psi$ plays an important role. 

\begin{theorem}\label{theorem:filtration}
    Given a model $\M = \ldiamondarg{S, \{R_i\}_{i\in \Ag}, V, Exp}$ and a formula $\varphi$, for any $\psi\in FL(\varphi)$ and $w\in\Sigma^\star$, 
    \begin{align*}
        \M|_w,s\vDash\psi \mbox{ iff } \M^\sim|_w, [s]\vDash\psi.
    \end{align*}
\end{theorem}

\begin{corollary}
    If $\varphi$ is POL-satisfiable, then $\varphi$ is satisfiable in a POL-model with at most $2^{|FL(\varphi)|}$ states.
    \label{corollary:smallmodelproperty}
\end{corollary}

\begin{proof}
    If $\modelM, s \models \varphi$ then $\modelM^\sim, [s] \models \varphi$. Note that $\modelM^\sim$ contains at most $2^{|FL(\varphi)|}$ states. 
\end{proof}

Although we have a small model property for $\POL$, it does not give an immediate decidability proof for the satisfiability problem of $\POL$. We note that in $\POL$ models, each state is associated with not only a valuation but also a regular expression. Given a satisfiable formula, the popular filtration technique \cite{modallogicblackburn} gives us a model where (i) the number of states is bounded above with respect to the size of Fischer-Ladner closure, and (ii) the number of possible valuations is bounded above due to the fact that the number of propositions is bounded by the size of the input formula. However, the size of the regular expression $\Exp(s)$ associated with each state $s$ may be arbitrarily large, and it is not straightforward to come up with a bound. We now provide a way to deal with this difficulty.

\section{Unraveling the filtrated model}
\label{sec:decidability}
\newcommand{\FBTS}{\textsf{BTS}\xspace}
\newcommand{\deltas}{\delta_s}
\newcommand{\Bbs}{{\Bb}_s}

We give alternative models to POL in terms of \emph{finite bubble transition structures} (\FBTS). They are {\em unraveled} POL models in which the expectation functions are represented with explicit transitions.
Furthermore, \FBTS's are syntactic in nature: 
epistemic relations are between states labeled by \emph{Hintikka sets} \cite{modallogicblackburn}.
In contrast to states in standard epistemic models, Hintikka sets also contain information about the future, e.g., if a Hintikka set contains $\ldiamondarg{a}\hat{K}_i p$, it says after $a$ is observed, $i$ should consider $p$ as possible.
%
In addition, 
these \FBTS's are always finite. Let us now describe them. 

\newcommand{\Tr}{\mathcal{T}}

\subsection{Finite Transition Model}

\textit{Hintikka sets} 
are sets of $\POL$ formulas satisfying some conditions, as defined below.

\newcommand{\hintikkaset}{H}
\begin{definition}[Hintikka set of Formulas]\label{defi:state}
    A Hintikka set~$\hintikkaset$ is a set of formulas satisfying following conditions: 
        \begin{enumerate}
            \item If $\psi$ does not start with negation, $\psi\in \hintikkaset$ iff $\neg\psi\notin \hintikkaset$.
            \item  $\psi_1\wedge\psi_2\in \hintikkaset$ iff $\{\psi_1, \psi_2\}\subseteq \hintikkaset$.
            \item  $\psi_1\vee\psi_2\in \hintikkaset$ iff $\psi_1\in \hintikkaset$ or $\psi_2\in \hintikkaset$.
            \item If $K_i\psi\in \hintikkaset$ then $\psi\in \hintikkaset$.
            \item If $\ldiamondarg{\pi_1 + \pi_2}\psi\in \hintikkaset$ then $\ldiamondarg{\pi_1}\psi\in \hintikkaset$ or $\ldiamondarg{\pi_2}\psi\in \hintikkaset$.
            \item  If $\ldiamondarg{\pi_1\pi_2}\psi\in \hintikkaset$ then $\ldiamondarg{\pi_1}\ldiamondarg{\pi_2}\psi\in \hintikkaset$.
            \item If $\ldiamondarg{\pi^\star}\psi\in \hintikkaset$ then either $\psi\in \hintikkaset$ or $\ldiamondarg{\pi}\ldiamondarg{\pi^\star}\psi\in \hintikkaset$.
            \item If $[\pi_1 + \pi_2]\psi\in \hintikkaset$ then $\{[\pi_1]\psi, [\pi_2]\psi\}\subseteq \hintikkaset$
            \item If $[\pi_1\pi_2]\psi\in \hintikkaset$ then $[\pi_1][\pi_2]\psi\in \hintikkaset$
            \item If $[\pi^\star]\psi\in \hintikkaset$ then $\{\psi, [\pi][\pi^\star]\psi\}\subseteq \hintikkaset$ 
        \end{enumerate}
\end{definition}
Besides the usual boolean conditions in \Cref{defi:state}, (4) corresponds to reflexivity of the knowledge relation. To satisfy diamond observation formulas $\ldiamondarg{\pi}\psi$, $\mathcal{L}(\pi)$ must contain a word. 
Thus, (5) corresponds to the word coming from $\pi_1$ or $\pi_2$, (6) corresponds to a word from $\pi_1$, followed by one from $\pi_2$, and (7) takes care of the iteration. In (8-10), the box formulas are considered in a dual manner.

\begin{example}
Consider $\phi := K_i(\ldiamondarg{a}(p\vee q)\wedge[a^\star]\ldiamondarg{a}(p\vee q))$. A Hintikka set containing $\phi$ is given as follows:
\begin{align*}
    H   = & \{\phi,
         \ldiamondarg{a}(p\vee q)\wedge[a^\star]\ldiamondarg{a}(p\vee q),\\
        & \ldiamondarg{a}(p\vee q), [a^\star]\ldiamondarg{a}(p\vee q),
        [a][a^\star]\ldiamondarg{a}(p\vee q)\}
\end{align*}
The second formula in H comes from (4), the third and fourth ones are due to (2), and the last one is a consequence of (10).
\end{example}

Let $\mathcal{K}_i(\hintikkaset) = \{K_i\psi\mid K_i\psi\in \hintikkaset\}$ be the set of knowledge formulas in $\hintikkaset$. 
We now define \textit{(epistemic) bubbles}. They are similar to $\POL$ models but differs in three ways. 
First, there is no expectation function $\Exp$ anymore. 
Second, the information about the future in a state $s$ previously stored in $\Exp(s)$ is now provided in the Hintikka set $L(s)$ attached to $s$. For instance if $\ldiaarg{a}\top \in L(s)$ it means that $a$ can be observed in $s$. 
Note that  all the necessary information about the expectation function and indistinguishability are now relational structures on Hintikka sets.
Third, a bubble is tailored for the corresponding formula $\phi$: Hintikka sets are given with respect to $\phi$ and the number of states is bounded by $2^{|FL(\phi)|}$ (as in~\Cref{corollary:smallmodelproperty}).




\begin{definition}\label{def:epibubble}
    A \emph{bubble} wrt a $\POL$ formula $\phi$ is a labelled relational structure $\ldiamondarg{S, \{R_i\}_{i\in \Ag}, L}$ such that: 
    \begin{enumerate}
        \item $S$ is a set of (abstract) states such that $0\leq|S|\leq 2^{|FL(\varphi)|}$
        \item $L:S\rightarrow 2^{FL(\varphi)}$ is a labelling function such that for every $s\in S$, $L(s)$ is a Hintikka set.

       \item $R_i\subseteq S\times S$, is a binary equivalence relation 
        such that:
        \begin{enumerate}
            \item For any $s\in S$, any formula $\hat{K_i}\psi\in L(s)$, there exists an $s'\in S$ such that $\psi\in L(s')$ and $(s,s')\in R_i$.

            \item For all $s', s''\in [s]_i$, the equivalence class of $s$ under $R_i$, $\mathcal{K}_i(L(s')) = \mathcal{K}_i(L(s''))$. 
        \end{enumerate}
    \end{enumerate}
\end{definition}

Point 3 describes the interaction between the indistinguishability relation $R_i$ and the Hintikka sets. If a state satisfies $\hat{K}_i\psi$, then there is an $i$-indistinguishable state satisfying $\psi$ (3a). Also, knowledge of agent $i$ (formulas of the form $K_i\psi$) 
are the same in all the $i$-indistinguishable states (3b).
We now introduce the notion of \emph{$a$-successor}. Considering a bubble $B$ that intuitively corresponds to a $\POL$ model $\M$,
an observation successor of $B$ is 
a bubble $B'$ 
which corresponds to $\M|_a$.

        




\begin{definition}
\label{defi:ObsSucc}
Let $B = \ldiamondarg{S, \{R_i\}_{i\in \Ag}, L}$ and $B' = \ldiamondarg{S', \{R'_i\}_{i\in \Ag}, L'}$ be two bubbles wrt to $\varphi$. Let $a\in\Sigma$. 
$B'$ is an $a$-\emph{successor} of $B$ ($B \xrightarrow{a} B'$) if the following conditions hold:    
\begin{enumerate}
        \item $S'\subseteq S$ and for all $s\in S'$, $L(s) \cap \setprop = L'(s) \cap \setprop$.

        \item 
        
        For all $s\in S, \ldiamondarg{a}\psi\in L(s)$ iff $(s\in S'$ and $\psi\in L'(s))$.


        \item For all  $s\in S'$, $[a]\psi\in L(s)$ iff $\psi\in L'(s)$.


        \item (Perfect Recall): $R_i' = R_i \cap (S' \times S')$.
\end{enumerate}
\end{definition}

Point 1 says that the set of states is decreasing when an observation $a$ is made, and that the valuations do not change (for the surviving states).
Point 2 says that any state $s$ satisfying a diamond formula $\ldiamondarg a \psi$  must survive after observing $a$, and then, must satisfy $\psi$. Point 3 says that if a state survives, same rule should apply for box formulas as well. Point 4 says that if an agent considers a state possible from the state $s$ in the projected (residuated) model, she should  consider it possible before the projection (Perfect Recall: $\ldiamondarg{a}\hat{K}_i\psi\rightarrow \hat{K}_i\ldiamondarg{a}\psi$).
Now we introduce the notion of \emph{bubble transition structure} (\FBTS)
of a formula $\varphi$, which is a deterministic automaton where nodes are bubbles.

\newcommand\nosuccessor\ddagger

\begin{definition}[\FBTS]
\label{defi:bubbletrans}
    Given a $\POL$ formula $\varphi$, a \emph{bubble transition structure} (\FBTS) of $\varphi$ is a 
    tuple
    $\B = \ldiamondarg{\Bb,\delta}$ where $\Bb$ is the (finite) set of all bubbles wrt $\phi$, and 
    $\delta: \Bb \times \Sigma \rightarrow \Bb \union \set{\nosuccessor}$
    is the 
    transition function 
   such that:
    \begin{enumerate}

        \item Some bubble $B^* = \ldiamondarg{S^*, \{R^*_i\}_{i\in \Ag}, L^*} \in \Bb$ 
        , called the \textit{initial bubble}, is
        s. th. there is an $s\in S^*$, with $\varphi\in L^*(s)$

        \item  For all $B = \ldiamondarg{S, \{R_i\}_{i\in \Ag}, L}\in\mathcal{B}$, either $\delta(B, a) = \nosuccessor$ or, $\delta(B, a) = B^a$ where 
        $B^a$ is an $a$-successor of~$B$.  

        \item For every $\ldiamondarg{\pi}\psi\in L(s)$ for any $s\in S$ of any node $B = \ldiamondarg{S, \{R_i\}_{i\in \Ag}, L}\in\mathcal{B}$ in $\B$, there is a $k$ length word $a_1a_2\ldots a_k\in\LL(\pi)$, a sequence of nodes (bubbles) $B^0 = B, B^{1},\ldots, B^{k}$, where each $B^{j}\in\mathcal{B}$, such that:
        \begin{enumerate}
        
            \item $B^{j} = \delta(B^{{j-1}}, a_{j})$, for all $1\leq j\leq k$.
            
            \item $s{\in }S^{k}$ and $\psi{\in} L^{k}(s)$, where $B_{k} {=} \ldiamondarg{S^{k}, \{R^{k}_i\}_{i\in \Ag}, L^{k}}$.
        \end{enumerate}
    \end{enumerate}
\end{definition}

In other words, a \FBTS can be thought of as a structure where each node represents some residue of the model represented by the bubble $B^*$ given in (1), which represents a model satisfying $\varphi$. (2) assumes that a model can have at most one residue structure for every letter $a\in\Sigma$. (3) says that if $\ldiamondarg{\pi}\psi$ is satisfied at a state in a model, then there exists a model which is residuated on some $w\in\LL(\pi)$ and the same state in the residuated model satisfies $\psi$. Note that a formula may have zero or multiple \FBTS's.

\begin{figure}
    \centering
\begin{center}
\scalebox{0.7}{
    \begin{tikzpicture}
        \node[bubble] (0) {
        \begin{tikzpicture}
        \node at (-2.7, 1.6) {$s$}; 
         \node[draw, text width=45mm] at (0, 1.6) (s) {$\phi$, 
         $[a]\bot$ \\
         $\hat{K_i}((\ldiamondarg{a}(p\vee q)\wedge[a^\star]\ldiamondarg{a}(p\vee q)$};

         \node at (-2.7, 0) {$t$};
           \node[draw, text width=45mm] (t) {
           $(\ldiamondarg{a}(p\vee q)\wedge[a^\star]\ldiamondarg{a}(p\vee q)$ \\
           $\ldiamondarg{a}(p \lor q)$, $[a^*]\ldiamondarg a (p \lor q)$, \\ $[a][a^*]\ldiamondarg a (p \lor q)$, \textcolor{blue}{$p$}};

           \draw[->] (s) edge[loop right, looseness=3] node[right] {$i$} (s); 
            \draw[->] (t) edge[ loop right, looseness=0.8] node[right] {$i$} (t); 

            \draw[<->] (s) edge node[right] {$i$} (t); 
        \end{tikzpicture}
        };

        \node[bubble] (1) at (6, 0) {
        \begin{tikzpicture}
            \node at (-1.6, 0) {$t$};
            \node[draw,text width=25mm] (t) {$p\lor q$, $p$, \\ $\ldiamondarg{a}(p \lor q)$,\\ $[a^*]\ldiamondarg a (p \lor q)$, \\$[a][a^*]\ldiamondarg a (p \lor q)$};
            \draw[->] (t) edge[loop right, looseness=1] node[right] {$i$}  (t);
           \end{tikzpicture}};
        \node[above = 0mm of 0]{$B^*$};
        \node[above = 0mm of 1]{$B$};
        \draw[obstransition] (0) edge node[above] {$a$} (1);
        \draw[obstransition]  (1) edge[loop right, looseness=1] node[right] {$a$} (1);
    \end{tikzpicture}}
\end{center}
    \caption{A \FBTS for $\varphi := [a]\bot \land \hat{K_i}(\ldiamondarg{a}(p\vee q)\wedge[a^\star]\ldiamondarg{a}(p\vee q))$. There are two bubbles: $B^*$ and $B$. There are two abstract states: $s$ and $t$. If a state appears in a bubble, it is labelled by a Hintikka set: for instance, $s$ in $B^*$ is labelled by $L^*(s) = \{\phi, [a]\bot, \dots \}$.}
    \label{fig:bubbletransitionstructure}
\end{figure}
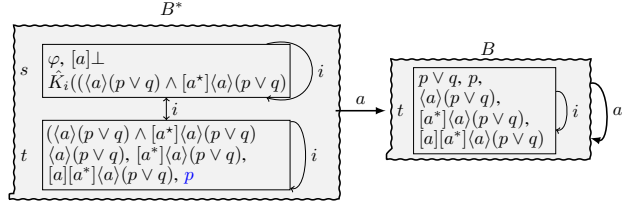

\begin{example}
Consider the \FBTS $\B$ in Figure~\ref{fig:bubbletransitionstructure},  where $\varphi := [a]\bot \land \hat{K_i}(\ldiamondarg{a}(p\vee q)\wedge[a^\star]\ldiamondarg{a}(p\vee q))$.

\begin{itemize}
    \item All formulas appearing in the labels are in $FL(\varphi)$ (Definition~\ref{defi:bubbletrans} (1))
    \item $\phi$ appears in the label of $s$ in $B^*$ (Definition~\ref{defi:bubbletrans} (2)).
    \item The bubble $B$ is an $a$-observation successor of $B^*$. Definition~\ref{defi:ObsSucc} (1): $S' \subseteq S$ is $\set{t} \subseteq \set{s, t}$ in our case. Definition~\ref{defi:ObsSucc} (2): $p$ appears both in $t$ in $B^*$ and in $t$ in $B$ (Definition~\ref{defi:bubbletrans} (3)).
\end{itemize}

By \Cref{defi:state}, since $[a^\star]\ldiamondarg{a}(p\vee q)\in L^*(t)$, $[a][a^\star]\ldiamondarg a (p \lor q)\in L^*(t)$. Now by Definition \textcolor{red}{\ref{defi:bubbletrans} (4)}, since $\ldiamondarg{a}(p\vee q)\in L^*(t)$, we have $(p\vee q)\in L(t)$. We also have $[a^\star]\ldiamondarg{a}(p\vee q)\in L^*(t)$, which again, by definition of Hintikka set gives rise to $\{\ldiamondarg{a}(p\vee q), [a][a^\star]\ldiamondarg{a}(p\vee q)\}\subseteq L^*(t)$.
\end{example}

\newcommand{\Ss}{\mathcal{S}}
\newcommand{\algcom}{\textbackslash\textbackslash}

\subsection{Completeness}

\begin{restatable}{theorem}{transmodelsound}
    \label{theorem:transmodelsound}
    If 
    $\varphi$ is satisfiable, then there is a $\FBTS$ of $\varphi$.
\end{restatable}
\noindent\textbf{Proof Sketch.} 
Suppose $\varphi$ satisfiable. By \Cref{corollary:smallmodelproperty}, there exists $\M$ with at most $2^{|FL(\phi)|}$ states and $s$ such that $\M, s \models \phi$.
We create a transition system $\B = \ldiamondarg{\Bb,\delta}$, where $\Bb$ contains exactly the bubbles $B^w$ corresponding to $\M|_w$ for some $w\in\Sigma^\star$.
The transition function 
$\delta$ is defined by
$$\delta(B, a) = \begin{cases}B^{wa} \text{, $\exists~ w$ such that $B = B^w$ and $\M|_{wa}$ exists} \\
\nosuccessor, \text{ otherwise}
\end{cases}$$

It can be shown that $\B$ is a $\FBTS$ of $\varphi$.

\newcommand{\pFTS}{\textsf{ITS}\xspace}


\subsection{Soundness}




\begin{restatable}{theorem}{transmodelcomplete}\label{theorem:transmodelcomplete}
If there is \FBTS of $\varphi,$ then $\varphi$ is satisfiable.
\end{restatable}

Consider a \FBTS $\B = \ldiaarg{\Bb,\delta}$ of $\phi$.
The proof of \Cref{theorem:transmodelcomplete} constitutes the construction of a pointed $\POL$ model $\M^\varphi, s_0$ satisfying $\phi$ out of $\B$. We construct the $\POL$ model $\M^\varphi = \ldiamondarg{S^\phi, \{R^\phi_i\}_{i\in \Ag}, V^\phi, \Exp}$ as follows:
 %
 $S^\phi = S^*$,  $R^\phi_i = R^*_i$ for all agents $i$,  $V^\phi(s) = L^*(s)\cap\mathcal{P}$, that is, the propositions of $\varphi$ that labels in $s$, for any $s\in S^*$.
 The state $s_0$ is some state such that $\phi \in L^*(s_0)$, where $B^* = \ldiamondarg{S^*, \{R^*_i\}_{i\in \Ag}, L^*}$ is the initial bubble.

 It remains to define the expectation function $\Exp$.
\newcommand{\Aut}{\ensuremath{\mathcal{A}}}
For each state $s$, we define $\Exp(s)$ as a regular expression that characterizes the language of the automaton $\Aut_s$ defined in the following:

\begin{itemize}
    \item[-] The set of states of $\Aut_s$ is $\Bb$;
    \item[-] The transition function of $\Aut_s$ is given by $\delta$;
    \item[-] The initial state of $\Aut_s$ is $B^*$;
    \item[-] The final states of $\Aut_s$ are the bubbles $B$ containing $s$.
\end{itemize}

As a state $s$ cannot resurrect, non-final states in $\Aut_s$ are absorbing: when a non-final state is reached by reading some word, it is impossible to reach a final state again.

\begin{example}
    Consider Figure~\ref{fig:bubbletransitionstructure}. In $\Aut_t$, both $B^*$ and $B$ are final. We set $\Exp(t) = aa^*$. In $\Aut_s$, only $B^*$ is final. We set $\Exp(s) = \epsilon$.
\end{example}

    \noindent 

The proof of Theorem~\ref{theorem:transmodelcomplete} ends with the following claim, implying that $\phi$ is satisfiable.
\begin{claim}
    $\modelM^\phi, s_0 \models \phi$.
\end{claim}
\begin{proof}
    We prove a more general result. We prove the following property $\mathcal{P}(\psi)$ by inducting on $\psi\in FL(\phi)$:
    \begin{align*}
        \mathcal{P}(\psi): & \begin{minipage}{7cm} for all bubbles $B\in\Bb$, for all states $s\in B$, if $\psi\in L(s)$ then
        $\modelM^\phi|_w,s\vDash\psi$
        for all words $w\in\Sigma^\star$ such that $B^*\xrightarrow{w}^* B$
        \end{minipage}
    \end{align*}
where $B^*\xrightarrow{w}^* B$ means that there is a path from $B^*$ to $B$ by reading the word $w$ in $\B$.

    \noindent {\bf Base Case.} The case $\psi = p$ is implied by the construction and point 1 of \Cref{defi:ObsSucc}.

    \medskip
    
    \noindent {\bf Inductive Step.} 
    Based on the syntax of $\psi$:
    \begin{itemize}
        \item $\psi = \hat{K}_i\chi$. As $\psi \in L(s)$, by \Cref{def:epibubble}, there is some $t$ such that $(s,t)\in R_i$ and $\chi\in L(t)$. By IH, $\modelM^\phi|_w,t\vDash\chi$. Due to point 4 (Perfect Recall) of \Cref{defi:ObsSucc}, the relation $(s,t)\in R_i$ is retained from $B^*$ along the path $B^*\xrightarrow{w}^* B$. Consider the construction of $\Aut_s$. Since all bubbles having $s$ in it are marked final, hence $Exp(s)\regdiv w\neq\emptyset$ since $w\in\LL(Exp( s))$. Hence by construction and the previous argument, the relation is retained in $\modelM^\phi$ as well since $s$ is still retained in $\modelM^\phi|_w$. Therefore $\modelM^\phi,s\vDash\hat{K}_i\chi$.

        \item $\psi = \ldiamondarg{\pi}\chi$. By point 3 of \Cref{defi:bubbletrans}, there is some $w'\in\Sigma^\star$ such that $\chi\in L'(s)$, where $L'$ is the labelling function for a bubble $B'$ and $B\xrightarrow{w'}^* B'$. Hence by IH, $\modelM^\phi|_{ww'},s\vDash\chi$. This implies $\modelM^\phi|_w,s\vDash\ldiaarg{\pi}\chi$. 
    \end{itemize}
    Hence the property $\mathcal{P}(\psi)$ is proved for any formula $\psi\in FL(\phi)$ for any state $s$ of any bubble $B$ in \FBTS. In particular, $\mathcal{P}(\phi)$ implies $\modelM^\phi,s_0\vDash\phi$ since $\phi\in L^*(s_0)$ of $B^*$.
\end{proof}

\section{POL Satisfiability by DPDL}
\label{sec:reductiontodpdl}
\newcommand{\translationsem}{sem}
\newcommand{\labelworld}{\ell}
\newcommand{\labelworldinitial}{\ell_0}
\newcommand{\setlabels[1]}{\mathfrak L_{#1}}
\newcommand{\proplabelsatisfiesformula}[2]
{@#1.#2}
\newcommand{\proprel}[3]
{R_{#1}(#2, #3)}
\newcommand{\propsurv}[1]
{surv(#1)}

Finally, to provide an algorithm for checking satisfiability of $\POL$, we now provide a translation of $\POL$ formulas to Deterministic Propositional Dynamic Logic ($\DPDL$) formulas and use the complexity results of $\DPDL$ \cite{DPDL-SAT-Ben-AriHP82}. Before proceeding further, let us have a brief look at the syntax and semantics of $\DPDL.$

\subsection{On $\DPDL$}

%
	Given a countable set of atomic propositions $\cP$, and a finite set of actions $\Sigma,$ the language of $\DPDL$ is given by the following:   
$$  \varphi := \top\ \  |\ \  p\in\cP\ \  |\ \  \varphi\vee\psi\ \ |\ \ \neg\varphi\ \ |\ \ \ldiamondarg{\pi}\psi,$$
where $\pi$ is regular expression over $\Sigma$.

\newcommand{\modelDPDL}{M}
	A $\DPDL$ model is given by: $\modelDPDL = \ldiamondarg{W,\{\rightarrow_a\}_{a\in\Sigma}, V},$ where, $W$ is a finite set of states, $\rightarrow_a\subseteq W\times W$ is a binary relation such that for every $a\in\Sigma$, and for every $w\in W$, if $(w,w_1)\in\rightarrow_a$ and $(w,w_2)\in\rightarrow_a$ then $w_1 = w_2$, and $V:W\rightarrow 2^{\cP}$ is the valuation function.

We extend the relation $\rightarrow_a$ to $\rightarrow_\pi$ for any general regular expression $\pi$ in the usual way:
		\begin{itemize}
			\item $\rightarrow_{\pi_1 + \pi_2} = \; \rightarrow_{\pi_1}\union\rightarrow_{\pi_2}$
			\item $\rightarrow_{\pi_1;\pi_2} = \!\{\!(w,v)\!\mid\!\exists u\in W\!:\! (w,u)\in\rightarrow_{\pi_1} 
             \!\wedge(u,v)\in\rightarrow_{\pi_2}\!\}$
			\item $\rightarrow_{\pi^\star} = \bigcup_{k\geq 0}\rightarrow_{\pi^k}$ where $\pi^k = \underbrace{\pi;\pi;\ldots;\pi}_{k \text{ times}}$ 
		\end{itemize}

	Given a DPDL model $\modelDPDL = \ldiamondarg{W,\{\rightarrow_a\}_{a\in\Sigma}, V}$ and a DPDL formula $\varphi$, we define $\MD,s\vDash\varphi$ for some $s\in W$ as usual by induction on $\phi$. We mention the modal case below that involves the language of the regular expression.
	\begin{itemize}
		\item $\MD,s\vDash \ldiamondarg{\pi}\psi$ iff there is a word $w\in\LL(\pi)$ such that $s\rightarrow_w t$ and $\MD,t\vDash\phi$
	\end{itemize}

To clarify the distinction between $\DPDL$ and $\PDL$ \cite{hareldynamiclogic}, let us consider the formula, $\ldiamondarg{a} p \land \ldiamondarg{a} \neg p.$ The formula is satisfiable in $\PDL,$ but not in $\DPDL.$

\subsection{Translation of POL into DPDL}

We introduce a translation from $\POL$-formulas into $\DPDL$-formulas. The idea is as follows. On the one hand, each observation operator~$\ldiaarg{\pi}$ is directly simulated by its $\DPDL$-dynamic operator counterpart. On the other hand, we encode each epistemic structure as a propositional theory.

Thanks to the filtration result (see \Cref{theorem:filtration}), we know that a satisfiable formula $\phi$ has a POL model with at most $2^{|FL(\phi)|}$ worlds. We can then pinpoint the worlds by labels $\labelworld, \labelworld', \dots$ from the set of labels $\setlabels^\phi = \set{1,\dots, 2^{|FL(\phi)|}}$.

We introduce special atomic formulas of the form $\proplabelsatisfiesformula\labelworld\psi$
whose intuitive meaning is 
`subformula $\psi$ is true in the $\labelworld$-th world'. We also introduce atomic propositions
$\proprel i \labelworld{\labelworld^\prime}$
whose intuitive meaning is `the $\labelworld$-th world is linked to the the $\labelworld'$-th world by the relation $R^i$'.
In addition, we introduce atomic proposition $\propsurv\labelworld$ that intuitively says that the $\labelworld$-th world has survived so far (meaning that the $\labelworld$-th world is still compatible with the observations that have been seen so far).



  Given a $\POL$-formula $\varphi$, we define 
  a $\DPDL$-formula~$\translationsem(\varphi)$ that encodes the semantics of $\varphi$: 
    \begin{itemize}
        \item $\translationsem(p) = \bigwedge_{l \in \setlabels^\phi} (@\labelworld.p\leftrightarrow \neg @\labelworld.\neg p)$
        \item $\translationsem(\neg\psi) = \bigwedge_{l \in \setlabels^\phi} (@\labelworld.\neg\psi\leftrightarrow \neg @\labelworld.\psi)$
        \item $\translationsem(\psi\vee\psi') = \bigwedge_{l \in \setlabels^\phi} (@\labelworld.(\psi\vee{\psi'})\leftrightarrow(@\labelworld.\psi\vee @\labelworld.{\psi'}))$
        \item $\translationsem(\hat{K}_i\psi) = \bigwedge_{l \in \setlabels^\phi} (@\labelworld.\hat{K}_i\psi\leftrightarrow
        (\bigvee_{l^\prime \in \setlabels^\phi}(\proprel i \labelworld {\labelworld^\prime}\wedge surv(\labelworld^\prime)\wedge @\labelworld^\prime.\psi)))$
        \item $\translationsem(\ldiamondarg{\pi}\psi) = \bigwedge_{l \in \setlabels^\phi} (@\labelworld.\ldiamondarg{\pi}\psi
        \leftrightarrow\ldiamondarg{\pi}(@\labelworld.\psi\wedge surv(\labelworld)))$
    \end{itemize}

Formula $\translationsem(\varphi)$ is about the `local' semantics of $\varphi$. For example, $\translationsem(\psi\vee\psi')$ explains the semantics of $\lor$ in the stage of $\psi\vee\psi'$; the semantics of $\psi$ and $\psi'$ are taken care of by $\translationsem(\chi)$ for the subformulas $\chi$ of $\psi$ and $\psi'$. Formula $\translationsem(\ldiamondarg{\pi}\psi)$ reflects the fact that the semantics of $\ldiamondarg{\pi}\psi$ is given by the DPDL-operator $\ldiamondarg{\pi}$ for a given world $\labelworld$ that should survive. 
Formula $\translationsem(\hat{K}_i\psi)$ expresses the semantics of $\hat{K}_i\psi$ (see \Cref{definition:POLtruthconditions}) in the propositional theory: 
the existence of a state is replaced by the disjunction \textcolor{red}{$\bigvee_{l^\prime \in \setlabels\phi}$}, the relation constraint is replaced by the propositional formula $\proprel i \labelworld {\labelworld^\prime}\wedge surv(\labelworld^\prime)$, and the truth of $\psi$ in the possible world by the proposition $@\labelworld^\prime.\psi$.

   Given a $\POL$-formula $\varphi$, we also define the $\DPDL$-formula $\cS_\varphi$ as the conjunction of the following expressions: 
    \begin{enumerate}
        \item $\bigwedge_{\psi\in FL(\varphi)}[\Sigma^\star]\translationsem(\psi)$
        \item $\bigwedge_{\labelworld \in \setlabels^\phi}((@\labelworld.p\rightarrow[\Sigma^\star]@\labelworld.p)\wedge ((@\labelworld.\neg p\rightarrow[\Sigma^\star]@\labelworld.\neg p)))$
        \item $\bigwedge_{\labelworld, \labelworld'}((\proprel i \labelworld {\labelworld^\prime}\rightarrow [\Sigma^\star]\proprel i \labelworld {\labelworld^\prime})
        \wedge (\lnot \proprel i \labelworld {\labelworld^\prime})  \rightarrow [\Sigma^\star] \lnot \proprel i \labelworld {\labelworld^\prime})$ 
        \item $\bigwedge_{i\in Agt}\bigwedge_{\labelworld\in\setlabels^\phi} \proprel i \labelworld \labelworld$
        \item $\bigwedge_{i\in Agt}\bigwedge_{\labelworld\in\setlabels^\phi} \bigwedge_{\labelworld'\in\setlabels^\phi} (\proprel i \labelworld {\labelworld'} \rightarrow \proprel i {\labelworld'} {\labelworld})$
       \item $\bigwedge_{i\in Agt}\bigwedge_{\labelworld, \labelworld', \labelworld''\in\setlabels^\phi}  ((
       \proprel i \labelworld {\labelworld'} \land \proprel i {\labelworld'} {\labelworld''} \rightarrow \proprel i {\labelworld} {\labelworld''})$
        \item $[\Sigma^\star]\bigwedge_{\labelworld}\bigwedge_{a\in\Sigma}\ldiaarg{a}\top$
        \item $[\Sigma^\star]\bigwedge_{\labelworld}\bigwedge_{a\in\Sigma}(\neg surv(\labelworld)\rightarrow\ldiaarg{a}\neg surv(\labelworld))$
    \end{enumerate}

Point 1 says that for all subformulas $\psi$, the semantics of $\psi$ is enforced anywhere in the DPDL model. Point 2 says that the truth value of any atomic proposition $p$ does not change at a given label/state $\labelworld$ when observations are made.
Point 3 says that the epistemic relations do not change. Note that labels $\labelworld, \labelworld'$ may not survive, that is handled by the proposition $surv(\labelworld)$. Points 4, 5, and 6 encode, respectively, the reflexivity, symmetry, and transitivity of the epistemic relation. Point~7 says that an $a$-successor always exists (even if no labels survive! The surviving mechanism is fully handled by propositions $surv(\labelworld)$). Point 8 says that a label $\labelworld$ that is dead remains dead.


\begin{definition}
    [Translation from POL to DPDL]
    Given a POL-formula $\phi$, we define $tr(\phi) := surv(\labelworldinitial)\wedge @\labelworldinitial.\phi \land\cS_\varphi$.
\end{definition}

In the translation $tr(\phi)$ above, we say that the state $\labelworldinitial$ must have survived (the empty list of observations so far), formula $\phi$  should be true in $\labelworldinitial$, and~$\cS_\varphi$ forces the semantics to be well-behaved.

\begin{proposition}\label{prop:dpdltransexpo}
Given $\POL$ formula $\phi$, $tr(\phi)$ is computable in exponential time in the size of $\phi$.
\end{proposition}
\begin{proof}
    Given a $\POL$ formula $\phi$, there are at most $O(|\phi|)$ many formulas in $FL(\phi)$, and by filtration 
    (\Cref{theorem:filtration}) there can be at most $O(2^{|\phi|})$ many unique labels. Therefore the time taken and the size of the formula $tr(\phi)$ is at most $O(|\phi|\times 2^{|\phi|})$.
\end{proof}

\begin{restatable}{proposition}{dpdltranscorrect}\label{prop:dpdltranscorrect}
$\phi$ is $\POL$-satisfiable iff $tr(\phi)$ is DPDL-satisfiable.
\end{restatable}

From Propositions~\ref{prop:dpdltransexpo} and~\ref{prop:dpdltranscorrect}, and from the fact that $\DPDL$ satisfiability is $\EXPTIME$-complete~\cite{DPDL-SAT-Ben-AriHP82} we get:

\begin{theorem}
    \label{theorem:upperbound}
    $\POL$-satisfiability is in $2\EXPTIME$.
\end{theorem}

\begin{proof}
    Here is a double-exponential algorithm for testing the satisfiability of $\phi$ in $\POL$:
    \begin{itemize}
        \item compute $tr(\phi)$ (exponential-time in $\phi$)
        \item test whether $tr(\phi)$ is $\DPDL$-satisfiable (exponential time in $tr(\phi)$, so double-exponential time in $\phi$).
    \end{itemize}
The result follows from the algorithm above.
\end{proof}



\section{Lower bound}
\label{sec:lowerbound}

In the following we provide a lower bound for the satisfiability problem of $\POL,$ and show that:

\begin{theorem}\label{final}
    $\POL$ satisfiability 
    is $2\EXPTIME$-complete. 
    \label{theorem:lowerbound}
\end{theorem}


The upper bound was shown in \Cref{theorem:upperbound}.
To prove $2\EXPTIME$-hardness in \Cref{theorem:lowerbound}, consider any $2\EXPTIME$ problem~$\anyproblem$. As $\AEXPSPACE = 2\EXPTIME$ \cite{DBLP:journals/jacm/ChandraKS81}, there is an alternating Turing machine $\tm$ deciding $\anyproblem$ in exponential space $e(|x|)$ where $x$ is the input, and $|x|$ is its length/size. The accepting and rejecting state are respectively denoted by $\qacc$ and $\qrej$.

We represent a configuration as a word of symbols. A \emph{symbol} can be either a letter (0, 1, or $\text{\textvisiblespace}$) written on the tape, or a pair $qa$ where $q$ is a state of $M$ and $a$ is a letter, or a special symbol $\leftmostrightmostsymbol$. 
The set of symbols is denoted by $\setsymbols$. We suppose that the state symbol precedes the letter on which the head is. All configuration words start and finish by the special symbol $\leftmostrightmostsymbol$. For instance, the configuration
%
\scalebox{0.6}{
    \begin{tikzpicture}[scale=0.6]
   \tikzstyle{cell}=[draw, minimum width=6mm, minimum height=6mm];
  \node[cell] at (5, 0) {0};
  \node[cell] at (4, 0) {1};
   \node[cell] (current) at (3, 0) {1};
   \node[cell] at (2, 0) {1};
   \node[cell] at (1, 0) {0};
   \node[cell] at (6, 0) {0};
   \node[cell] at (7, 0) {\textvisiblespace};
 \node[cell] at (8, 0) {\textvisiblespace};
  \node[inner sep=0.5mm] (q) at (3, 1.5) {$q$};
  \draw[-latex] (q) -- (current);
\end{tikzpicture}}
%
where the tape contains $011100\textvisiblespace\textvisiblespace\dots$, the machine is in state $q$, and the head is under the third left-most cell
is represented by the word 
\scalebox{0.6}{
    \begin{tikzpicture}[scale=0.6]
   \tikzstyle{cell}=[draw, minimum width=6mm, minimum height=6mm];
   \node[cell] at (0, 0) {$\leftmostrightmostsymbol$};
   \node[cell] at (1, 0) {0};
   \node[cell] at (2, 0) {1};
   \node[cell] (current) at (3, 0) {q1};
   \node[cell] at (4, 0) {1};
   \node[cell] at (5, 0) {0};

   \node[cell] at (6, 0) {0};
    \node[cell] at (7, 0) {\textvisiblespace};
      \node[cell] at (8, 0) {\textvisiblespace};
   \node[cell] at (9, 0) {$\leftmostrightmostsymbol$};

\end{tikzpicture}}
.

Without loss of generality (w.l.o.g.), we suppose that the machine $M$ switches between universal and existential states and starts with an existential state. 
Also, w.l.o.g. each configuration has at most two successor configurations. So we use two functions $\successorfunctiona$ and $\successorfunctionb$ such that given three consecutive symbols $\symbolfirstconfig \symbolsecondconfig \symbolthirdconfig \in \setsymbols^3$, $\successorfunctiona(\symbolfirstconfig \symbolsecondconfig \symbolthirdconfig)$ (resp.
$\successorfunctionb(\symbolfirstconfig \symbolsecondconfig \symbolthirdconfig)$) is the symbol in the middle position after the first transition (resp. the second transition), see Figure~\ref{figure:tm-successorfunction}.
Taking a transition is modeled by the program $a \cup b$.
Figure~\ref{fig:hardnessbigpicture} explains the idea behind the reduction, namely how to represent a computation tree inside a $\POL$ model.

\tikzstyle{cell}=[draw, minimum width=7mm, minimum height=6mm]
\tikzstyle{threecells} = [decoration={
        brace, mirror, raise=-0.6mm}, decorate, line width=0.3mm]
 \newcommand{\yred}{-0.8}
 \newcommand{\ygreen}{-1.1}
 \newcommand{\yblue}{-0.95}
 \tikzstyle{gives} = [line width=0.3mm, -latex]

\newcommand{\figuretransition}[4]{
\scalebox{0.7}{
   \begin{tikzpicture}[xscale=0.7, yscale=0.5]
   \node[cell] at (0, 0) {};
   \node[cell] at (5, 0) {};
   \node[cell] at (4, 0) {$#3$};
   \node[cell] at (3, 0) {$#2$};
   \node[cell] at (2, 0) {$#1$};
   \node[cell] at (1, 0) {};
   \node[cell] at (6, 0) {};
   
\node[cell] at (0, -2.5) {};
   \node[cell] at (5, -2.5) {};
   \node[cell] at (4, -2.5) {};
   \node[cell] at (3, -2.5) {$#4$};
   \node[cell] at (2, -2.5) {};
   \node[cell] at (1, -2.5) {};
   \node[cell] at (6, -2.5) {};

\draw[green!50!black, threecells] (1.55, \yred) -- (4.45, \yred);
   \draw[green!50!black, gives] (3, \yred) -- (3, -1.9);
  
\end{tikzpicture}
}
}

\begin{figure}
    \centering
    \begin{subfigure}{0.25\textwidth}
        \figuretransition{\symbolfirstconfig}{\symbolsecondconfig}{\symbolthirdconfig}?
    \end{subfigure}
   \begin{subfigure}{0.20\textwidth}
       \scalebox{0.7}{
       \begin{tikzpicture}[xscale=0.7, yscale=0.5]
   \node[cell] at (0, 0) {};
   \node[cell] at (1, 0) {3};
   \node[cell] at (2, 0) {$q0$};
   \node[cell] at (3, 0) {$2$};
   \node[cell] at (4, 0) {$1$};
   \node[cell] at (5, 0) {$0$};
   \node[cell] at (6, 0) {};

   \draw[red, threecells] (0.6, \yred) -- (3.4, \yred);
   \draw[green!50!black, threecells] (1.55, \ygreen) -- (4.45, \ygreen);
   \draw[blue, threecells] (2.6, \yblue) -- (5.4, \yblue);
   
\node[cell] at (0, -2.5) {};
   \node[cell] at (1, -2.5) {};
   \node[cell, text=red] at (2, -2.5) {1};
   \node[cell, text=green!50!black] at (3, -2.5) {q'2};
   \node[cell, text=blue] at (4, -2.5) {1};
   \node[cell] at (5, -2.5) {};
   \node[cell] at (6, -2.5) {};

  \draw[red, gives] (2, \yred) -- (2, -1.9);
   \draw[green!50!black, gives] (3, \ygreen) -- (3, -1.9);
  \draw[blue, gives] (4, \yblue) -- (4, -1.9);
\end{tikzpicture}}
   \end{subfigure}
    \caption{Successor function in a Turing machine: from three consecutive symbols $\alpha\beta\gamma$ the successor function tells the symbol written in the middle~(?). On the right, we give an example of a $a$-transition `if the machine is in state $q$ with 0 under the head, then write 1 and move to right and go into state $q'$'. In particular, we have $\successorfunctiona(3~q0~2)=1, \successorfunctiona(q0~2~1)=q'$ and $\successorfunctiona(021)=2$. \label{figure:tm-successorfunction-example}\label{figure:tm-successorfunction}}
\end{figure}
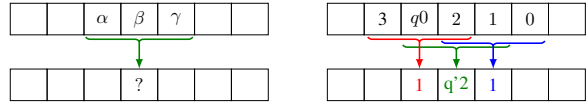

\textbf{Encoding a configuration. }
    We encode a superposition of \emph{three} configurations into an epistemic structure, and then say that they are equal.

   As there are two agents, thus two epistemic modalities $K_i$ and $K_j$ we
can simulate a standard K modal logic $\lbox$. For the rest of
the proof, we consider such a modality $\lbox$ and its dual~$\ldia$. We also introduce $\lbox^{k} \phi$ for $\lbox \dots \lbox \phi$ where $\lbox$ is repeated $k$ times, and $\lbox^{\leq k} \phi$ for $\phi \land \lbox \phi \land \dots \land \lbox^k \phi$.

    We create a formula that ensures the existence of a binary tree in the epistemic structure. A \emph{position} in a word is a number in $\set{0, 1, \dots, e(|x|)}$. Each leaf is tagged with a 3-tuple of positions $\positionfirstconfig, \positionsecondconfig, \positionthirdconfig$ in the tape, and a 3-tuple of symbols $\alpha, \beta, \gamma$, written respectively at position $\positionfirstconfig, j, k$,  in respectively the first, second and third configuration -- the values of  $\positionfirstconfig$, $\positionsecondconfig$, $k$ by the truth values of propositions $p_1, \dots, p_n$, $p_{n+1}, \dots, p_{2n}$
    and $p_{2n+1}, \dots, p_{3n}$, respectively.     A position in the tape is a number between 0 and $e(|x|) - 1$. As $e(|x|)$ is exponential in $|x|$, the integer $n$ above is polynomial in $|x|$. Our binary tree branches over the values for atomic propositions $p_1, \dots, p_{3n}$. To do so, we use the modal logic formula given in \cite{modallogicblackburn} and already used in \cite{krpolsf23}:
\newcommand{\bitindex}{m}
\begin{equation}\bigwedge_{\ell<6n}\!\!\!\lbox^\ell\!\! \left(\!\!\ldia p_\ell \land \ldia \lnot p_\ell \land \bigwedge_{\bitindex<\ell} 
\begin{array}{l}
(p_\bitindex {\limply} \lbox p_\bitindex) \land \\ 
(\lnot p_\bitindex {\limply} \lbox \lnot p_\bitindex)
\end{array}
\!\!\right)
\end{equation}
%
 %
%
   In order to 
   select some positions, we make the values of 
   propositions $p_\bitindex$ `observable'. To do that we introduce new observation symbol $p_\bitindex, \bar p_\bitindex \in \Sigma$ and the 
   constraints:
\begin{align}
  [\transitionsprogram]  \square^{3n}[\Sigma^*](p_\bitindex \lequiv \ldiaarg{p_\bitindex}\top \land [\bar p_\bitindex]\bot)
  \\
  [\transitionsprogram] \square^{3n}[\Sigma^*](\lnot p_\bitindex\lequiv \ldiaarg{\bar p_\bitindex}\top \land [p_\bitindex]\bot)
\end{align}

In other words, being able to observe $p_\bitindex \in \Sigma$ (resp. $\bar p_\bitindex \in \Sigma$) means that $p_\bitindex$ is true (resp. false).

For each symbol $\alpha$, we introduce observations $1{:}\alpha$, $2{:}\alpha$, $3{:}\alpha$. They are observable when symbol $\alpha$ is written in the current cell of respectively the first, second, and third configuration.
The following three formulas say that there is a unique symbol written at each position $\positionfirstconfig$, $\positionsecondconfig$, $\positionthirdconfig$:
\begin{align}
[\transitionsprogram]\square^{3n} \bigxor_{\alpha \in \setsymbols} \formulasymboliconfig \alpha
\end{align}
for $i=1..3$, and where $\bigxor$ is the XOR operator.

The two following formulas say that the symbols of the cells do not change when observing some positions:
\begin{align}
[\transitionsprogram] \square^{3n} (\bigwedge_{\alpha \in Symb} \formulasymboliconfig \alpha \nonumber \\
\limply  [(p_1\cup \bar{p_1}\cup  \dots \cup  p_{3n}\cup \bar p_{3n})^*]  \formulasymboliconfig \alpha)
\\
[\transitionsprogram] \square^{3n} (\bigwedge_{\alpha \in Symb} \lnot \formulasymboliconfig \alpha \nonumber \\
\limply  [(p_1\cup \bar{p_1}\cup  \dots \cup  p_{3n}\cup \bar p_{3n})^*]  \lnot \formulasymboliconfig \alpha)
\end{align}
%
%
We now say that for all $i=1..3$, all leafs with the same position $\positioniconfig$  contains the same symbol in the $i$-th configuration:
\begin{equation}
[\transitionsprogram][\text{choose $\positioniconfig$}]  \bigvee_{\alpha \in \setsymbols} \square^{3n} \formulasymboliconfig \alpha
\end{equation}
where \text{choose $\positionfirstconfig$}, \text{choose $\positionsecondconfig$}, \text{choose $\positionthirdconfig$} are respectively the programs $(p_1 \cup \bar p_1) \dots (p_n \cup \bar p_n)$, $(p_{n+1} \cup \bar p_{n+1}) \dots (p_{2n} \cup \bar p_{2n})$ and $(p_{2n+1} \cup \bar p_{2n+1}) \dots (p_{3n} \cup \bar p_{3n})$.

Finally we say that the \emph{three} configurations are equal. When two positions - say $\positionfirstconfig$ and $\positionsecondconfig$ - are equal, then the symbol located at $\positionfirstconfig$ in the first configuration and the symbol located at $\positionsecondconfig$ in the second configuration are equal.
More generally, for $i,j=1..3$, $i<j$:
\begin{equation}
[\transitionsprogram]\square^{3n} (\positioniconfig{=}\positionjconfig) \limply \bigvee_{\alpha \in \setsymbols} \formulasymboliconfig \alpha \land \formulasymboljconfig \alpha 
\end{equation}
where $\positionfirstconfig{=}\positionsecondconfig$ is 
a Boolean formula saying that $p_1, \dots, p_{n-1}$ and $p_{n+1}, \dots, p_{2n}$ encode the same number, and similar others.

\textbf{Presence of the complete binary tree after transitions. }
The following formula says that the existence of the $a$-transition ($a$ can be observed at the root) implies that $a$ can be observed at all nodes of the binary tree. Also if there is no $a$-transition, then $a$ is not observable at all nodes of the tree. Same for $b$. This is captured by the following scheme for $\psi$ being $\ldiaarg{a}\top$, $\lnot \ldiaarg{a}\top$, $\ldiaarg{b}\top$, $\lnot \ldiaarg{b}\top$:
\begin{align}
    [\transitionsprogram](\psi \rightarrow \lbox^{\leq 6n} \psi)
 \end{align}

\textbf{Leftmost and rightmost cells. }
We impose that the leftmost and rightmost cells always contain symbol $\leftmostrightmostsymbol$.
\begin{align}
   [\transitionsprogram]\lbox^{3n}(\positionfirstconfig=0 \limply  \formulasymbolfirstconfig \leftmostrightmostsymbol) \\
    [\transitionsprogram]\lbox^{3n}(\positionfirstconfig=e(|x|)-1 \limply  \formulasymbolfirstconfig \leftmostrightmostsymbol)
\end{align}

\textbf{Initial configuration.}
 At position 1, there is the initial state $q_0$. At position 2 is the first letter $x_1$ of $x$. The last letter~$x_n$ is at position $n+1$.
\begin{align}
    \lbox^{3n} (\positionfirstconfig {=} 1) \limply \formulasymbolfirstconfig {q_0x_1} \land (\positionfirstconfig {=} 2) \limply \formulasymbolfirstconfig {x_2} \nonumber \\
    \dots \land (\positionfirstconfig {=} n) \limply \formulasymbolfirstconfig {x_n}
\end{align}

After the word, we have the blank symbol $\text{\textvisiblespace}$ :
\begin{equation}
    \lbox^{3n} (\positionfirstconfig \geq n+1 \land \positionfirstconfig < e(|x|)-1) \limply \formulasymbolfirstconfig {\text{\textvisiblespace}}
\end{equation}

\textbf{Transitions. } W.l.o.g. we suppose the machine stops after writing $\textvisiblespace$. We define the two formulas
 $\formulaaccepted := \ldia^{3n} \formulasymbolfirstconfig {\qacc\textvisiblespace} $ and $\formularejected := \ldia^{3n} \formulasymbolfirstconfig {\qrej\textvisiblespace}$ meaning that the current configuration is respectively an accepting or rejecting one.
    We set $\formulaterminal := \formulaaccepted \lor \formularejected$.
    
We encode $a$- and $b$-transitions as follows.  We focus on leaves containing consecutive cells: a cell at position $\positionfirstconfig$ in the 1st configuration, the cell at  $\positionfirstconfig+1$ in the 2nd configuration, and the cell at  $\positionfirstconfig+2$ in the 3rd configuration. The superposition of the three configurations helps us to have access to three symbols $\alpha$, $\beta$, $\gamma$ in consecutive cells. We then use $\successorfunctiona$ and $\successorfunctionb$ to get the next middle symbol (at position $\positionfirstconfig+1$ in the 2nd configuration). 
\begin{align}
\begin{array}{l}
    [\transitionsprogram](\lnot \formulaterminal \limply \lbox^{3n} \left(\begin{array}{c}\positionsecondconfig{=}\positionfirstconfig{+}1 \land \\\positionthirdconfig{=}\positionsecondconfig{+}1\end{array}\right) \limply 
    \\
    \bigwedge_{\symbolfirstconfig \symbolsecondconfig \symbolthirdconfig \in \setsymbols^3} (\formulasymbolfirstconfig \symbolfirstconfig
    \land \formulasymbolsecondconfig \symbolsecondconfig 
    \land \formulasymbolthirdconfig \symbolthirdconfig) \\
    \limply (\ldiaarg{a} \formulasymbolsecondconfig {\successorfunctiona(\symbolfirstconfig \symbolsecondconfig \symbolthirdconfig)} \land \ldiaarg{b} \formulasymbolsecondconfig {\successorfunctionb(\symbolfirstconfig \symbolsecondconfig \symbolthirdconfig)})
\end{array}
\end{align}

In a terminal configuration, the execution stops meaning that we do not take transitions anymore ($[\transitionprogram] \bot$).
\begin{align}
    [\transitionsprogram] (\formulaterminal \limply  [\transitionprogram] \bot)
\end{align}

\textbf{Universal and existential configurations.}
\newcommand{\propexistential}{\ldiaarg{\exists}\top}
We introduce an observation symbol $\exists$, which is observable iff the current configuration is existential. Existential and universal configurations are always alternating:
\begin{align}
  [\transitionsprogram](\propexistential \limply [\transitionprogram] \lnot \propexistential)
\\
  [\transitionsprogram] (\lnot \propexistential \limply [\transitionprogram] \propexistential)
\end{align}

\textbf{Winning condition.}
The two following formulas explain what winning means at the final configurations: 
\begin{align}
  [\transitionsprogram] (\formulaaccepted \rightarrow \propwin) \\
 [\transitionsprogram] (\formularejected \rightarrow \lnot \propwin)  
\end{align}
The two following formulas explain what winning means at a non-terminal configuration:
\begin{align}
 [\transitionsprogram] ((\lnot \formulaterminal  \land  \lnot \propexistential) \nonumber \\
 \limply (\propwin   \lequiv [\transitionprogram] \propwin)
\\
 [\transitionsprogram] ((\lnot \formulaterminal  \land  \propexistential) \nonumber \\
 \limply (\propwin   \lequiv\ldiaarg{\transitionprogram} \propwin)
\end{align}

The following formula says that the initial configuration should be winning and is existential:
\begin{align}
    \propwin \land \propexistential
    \label{equation:lowerboundlastequation}
\end{align}

We define $tr(x)$ to be the conjunction of formulas (1-\ref{equation:lowerboundlastequation}) and the two following propositions conclude the proof.

\begin{restatable}{proposition}{lowerboundtrpolytime}
    $tr(x)$ is computable in poly-time in~$|x|$.
\end{restatable}

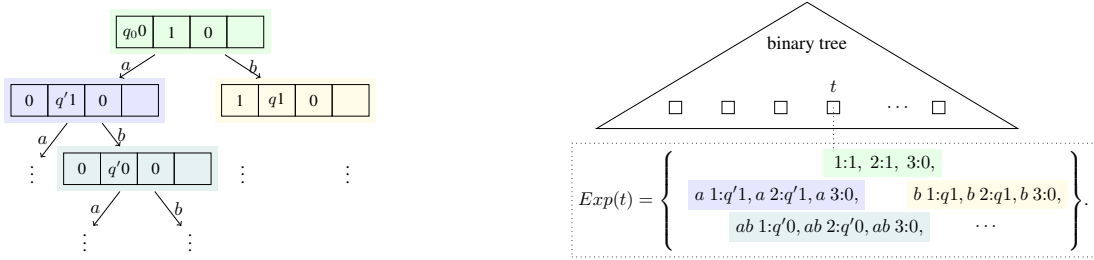
\begin{figure*}
    \centering
    \begin{subfigure}{0.46\textwidth}
    \centering
    \scalebox{0.7}{
    \begin{tikzpicture}[yscale=1.3]
        \node[fill=green!10!white] (root) {\configurationdraw {$q_0$0} 1 0};
        \node[fill=blue!10!white] (a) at (-2, -1) {\configurationdraw {0} {$q'$1} 0};
        \node[fill=blue!50!green!10!white] (ab) at (-1, -2) {\configurationdraw{0}{$q'$0}{0}};
        \node[fill=yellow!10!white] (b) at (2, -1) {\configurationdraw {1} {$q$1} 0};
        
        \node (aa) at (-3, -2) {$\vdots$};
        
        \node (ba) at (1, -2) {$\vdots$};
        \node (bb) at (3, -2) {$\vdots$};
        \node (aba) at (-2, -3) {$\vdots$};
        \node (abb) at (0, -3) {$\vdots$};
        \draw (root) edge[->] node[left] {$a$} (a) edge[->] node[right] {$b$} (b);
        \draw (a) edge[->] node[left] {$a$} (aa) edge[->] node[right] {$b$} (ab);
        \draw (ab) edge[->] node[left] {$a$} (aba) edge[->] node[right] {$b$} (abb);
    \end{tikzpicture}}
    \caption{Computation tree of $M$ on input $x = 01$. Each node of the three is a configuration of the machine. The root contains the initial configuration.
    The branching directions are called $a$ and $b$.
    }    
    \end{subfigure}
    \hfill
    \begin{subfigure}{0.5\textwidth}
    \centering
    \scalebox{0.7}{
    \begin{tikzpicture}[yscale=0.8]
        \draw (0, -1) -- (4, -4) -- (-4, -4) -- cycle;
        \node at (0, -2) {binary tree};
        \foreach \x in {-2.5, ..., 0.5} {
            \node[draw] at (\x, -3.5) {};
        }
        \node at (1.75, -3.5) {$\dots$};
        \node[draw] at (2.5, -3.5) {};
        \node  at (0.5, -3) {$\leaf$};
        \draw[dotted] (0.5, -3.5) -- (0.5, -4.5);

        \node[draw,dotted] at (0.5, -5.7) 
  {  $Exp(\leaf) = \left\{
    \raisebox{8mm}{
    \scalebox{1}{
    \newcommand{\examplethreeconfigs}[3]{\ensuremath{#1~\obssymbolfirstconfig {#2}, #1~\obssymbolsecondconfig {#2},  #1~\obssymbolthirdconfig {#3}, }}
    \begin{tikzpicture}[baseline=0mm, yscale=0.6]
        \node[fill=green!10!white] at (0, 0) {$
        \examplethreeconfigs{}10$};

        \node[fill=blue!10!white] at (-2, -1) {$\examplethreeconfigs a {q'1} {0}$};
        
         \node [fill=yellow!10!white] at (2, -1) {$\examplethreeconfigs b {q1}{0}$};
           \node[fill=blue!50!green!10!white] at (-1, -2) {$\examplethreeconfigs{ab}{q'0}{0}$};

        \node at (2, -2) {$\dots$};
    \end{tikzpicture}}}
    \right\}$.
    };no
    \end{tikzpicture}
    }
    \caption{The corresponding $\POL$ model. The epistemic structure encodes a binary tree. Each leaf $\leaf$ corresponds to three cell positions. The figure examplifies a leaf $\leaf$ where the first and second position are both the second cell while the third position is the third cell.}
    \end{subfigure}
    \caption{A computation tree and its corresponding $\POL$ model.}
    \label{fig:hardnessbigpicture}
\end{figure*}

\begin{restatable}{proposition}{lowerboundtrcorrect}
    $x$ is $\anyproblem$-positive iff $tr(x)$ is POL-satisfiable.
\end{restatable}

We thus reduce any $2\EXPTIME$-problem $\anyproblem$
to the satisfiability problem of $\POL$, proving the latter to be $2\EXPTIME$-hard, thus proving Theorem \ref{final}.


\section{Related work}
\label{section:relatedworks}
\newcommand{\CTLK}{\textsf{CTLK}}
\textbf{Propositional Dynamic Logics.} In $\POL$, a residuated model $\M|_w$ can have at most a unique successor $\M|_{wa}$ for a letter $a$ in the alphabet. This gives rise to a $\Sigma$-labelled transition structure equivalent to a model where each node is an epistemic skeleton of some $\M|_w$. 
Thus we get a natural connection with a Deterministic Propositional Dynamic Logic (DPDL) \cite{dpdlbenari82} model structure. But, there is a significant difference: each node in our constructed transition structure is an epistemic model, whereas in DPDL, it is a propositional valuation. To the best of our knowledge, no such deterministic structures involving epistemic constructions have been studied beforehand. 

Epistemic Propositional Dynamic Logic (EPDL) has been studied in \cite{epdlli18}. As in $\POL$, each node of an EPDL model is an epistemic structure, and it assumes \emph{perfect recall}. However, the transitions in EPDL are non-deterministic ones. Moreover, in EPDL, valuations may change when actions are executed, while that is not the case when observations are made in POL.


Another study on a PDL-like logic with epistemic operators \cite{heinemannpdlk07} concerns single agent knowledge formulas, where, verification of \textit{only} knowledge formulas in the regular expressions is considered. Although, they have given a hardness result (EXPTIME-hard) that follows from PDL, they do not have any matching upper bound. 

\noindent\textbf{Temporal Logics.} Since we are dealing with sequences of observations and how the expected observation expressions get residuated, there is a subtle temporal aspect to $\POL$. 
In $\LTLK$, the interaction between time and knowledge assumes \emph{perfect recall} and \emph{synchronous} rules, which also hold in $\POL$. 
But, there are some differences in the \textit{linearity} aspects. More precisely, since $\POL$ deals with multiple letters in the transition alphabet, knowledge does not change linearly, in the sense that, knowledge can change depending on the observations that occur. At a single point of occurrence this can be any one of the letters in the alphabet. We note here that a valuation in a state in the next temporal transition in $\LTLK$ may change, whereas such a valuation remains consistent in POL. Thus, $\neg p \land X p$ is satisfiable in $\LTLK$, whereas $\neg p \land \langle a\rangle p$ is not satisfiable in $\POL$, which leads to a difference in expressive power. From the complexity viewpoint, the satisfiability problem of $\LTLK$ is non-elementary, whereas for  $\POL$, it is $\mathsf{2EXPTIME}$-complete. It is interesting to note that both $\LTLK$ and $\POL$ can express that \emph{after some finite sequence of observations, $\varphi$ holds}, but only $\POL$ can express that \emph{after some even occurrences of observations, $\phi$ holds}. Resolving the intricate relationship between $\LTLK$ and $\POL$ needs further study.

Since knowledge updates \textit{branch out} in $\POL$ depending upon the action observed, Computational Tree Logic with epistemic operators ($\CTLK$) \cite{ctlkdima08} forms a close neighbour, whose satisfiability problem turns out to be undecidable. We note here that unravelled $\POL$ models create indistinguishability within the nodes (distinct $\POL$ models) themselves, whereas, indistinguishability in a $\CTLK$ model occurs among nodes across the tree. 


\section{Perspectives}
\label{section:conclusion}
Our main contribution has been to show that the satisfiability problem of $\POL$ (with Kleene star) is 2EXPTIME-complete. Now, we plan to study more tractable fragments of $\POL$, and also more expressive extensions of $\POL$, for instance, when expectations are context-free grammars.

A dynamic extension of $\POL$ mentioned in \cite{vanhidden2014}, Epistemic Protocol Logic ($\EPL$) is yet to be studied from the computational viewpoint. This logic is similar to $\DEL$: $\EPL$ also comes with operators that have pointed event or action model like $\DEL$, which \textit{assigns} expectations to the $\POL$ models. Such action models deal with $\PDL$-like regular expressions with Boolean verifiers. Complexity studies for the logic are open problems.

\paragraph{Acknowledgement. } This work was supported by the ANR EpiRL project ANR-22-CE23-0029. 
\bibliographystyle{unsrt}  
\bibliography{sample}

\newpage
\appendix

\section{Proofs from Section~\ref{sec:finitemodelproperty}}

{\bf Statement} The filtrated relation $R^\sim_i$ is symmetric.

\begin{proof}
    Let us assume $([s], [s'])\in R^\sim_i$. Hence
    \begin{itemize}
        \item {\bf Condition 1.} Since there exists $s_1\in [s]$ and $s_2\in [s']$ such that $(s_1, s_2)\in R_i$ and $R_i$ is symmetric, hence $(s_2, s_1)\in R_i$.

        \item {\bf Condition 2.} Take a general $\hat{K}_i\psi\in FL(\varphi)$. We assume that $\M,s\vDash\psi\vee\hat{K}_i\psi$. This implies $\M,s\vDash\psi$ or $\M,s\vDash\hat{K}_i\psi$.
        \begin{itemize}
            \item {\bf Case 1.} If $\M,s\vDash\psi$ and since $(s, s')\in R_i$, hence $\M,s'\vDash\hat{K}_i\psi$.

            \item {\bf Case 2.} If $\M,s\vDash\hat{K}_i\psi$. Hence there is a $t$ such that $(s, t)\in R_i$ and $\M,t\vDash\psi$. Since $(s, s')\in R_i$ as well and $R_i$ is equivalence relation, therefore $(s', t)\in R_i$ which further implies $\M,s'\vDash\hat{K}_i\psi$.
        \end{itemize}
    \end{itemize}

    Since both the condition satisfies for $([s'], [s])\in R^\sim_i$ hence the result.
\end{proof}

\expfinitemodel*
\begin{proof}
        We prove that using induction on $\pi$.

        \noindent\textbf{Base Case $\pi = a$}. Consider $\M,s\vDash\ldiamondarg{a}\psi$. Hence $s$ is in $\M|_a,s\vDash\psi$. Since $s\sim s'$, $\M,s'\vDash\ldiamondarg{a}\psi$ and hence follows.

        \noindent\textbf{Inductive case.}
        \begin{itemize}
            \item $\pi = \pi_1 + \pi_2$. $\M,s\vDash\ldiamondarg{\pi_1 + \pi_2}\psi$ hence there is a $w\in\LL(\pi+\pi_2)$ and $\M|_w,s\vDash\psi$. Now since $\M,s\vDash\ldiamondarg{\pi_1 + \pi_2}\psi$, hence $\ldiamondarg{\pi_1}\psi$ or $\ldiamondarg{\pi_2}\psi$ is satisfied in $s$. Wlog, suppose $\M,s\vDash\ldiamondarg{\pi_1}\psi$, hence $\M,s'\vDash\ldiamondarg{\pi_1}\psi$ by definition of $s\sim s'$. And hence by IH, our claim holds.

            \item $\pi = \pi_1\pi_2$. $\M,s\vDash\ldiamondarg{\pi_1\pi_2}\psi$, hence $\M,s\vDash\ldiamondarg{\pi_1}\ldiamondarg{\pi_2}\psi$. hence there is a $w\in\LL(\pi_1)$ such that $s$ survives in $\M|_w$ and $\M|_w,s\vDash\ldiamondarg{\pi_2}\psi$. Therefore, by IH $\M|_w,s'\vDash\ldiamondarg{\pi_2}\psi$ and $s'$ survives in $\M|_w$. 

            \item $\pi = \pi_1^\star$. $\M,s\vDash\ldiamondarg{\pi_1^\star}\psi$. This can only happen iff there is a word $w\in\LL(\pi_1^k)$ for some $k\geq 0$ such that $\M|_{w},s\vDash\psi$. Assume the $w = a_1a_2\ldots a_m$. Hence, $\M,s\vDash\ldiamondarg{a_1}\ldiamondarg{a_2}\ldots\ldiamondarg{a_m}\psi$ and $\ldiamondarg{a_1}\ldiamondarg{a_2}\ldots\ldiamondarg{a_m}\psi\in FL(\varphi)$, which means $\M,s'\vDash\ldiamondarg{a_1}\ldiamondarg{a_2}\ldots\ldiamondarg{a_m}\psi$ which gives the result.
        \end{itemize}
        This completes the proof.
    \end{proof}

\section{Proofs from Section~\ref{sec:decidability}}
\transmodelsound*
\begin{proof}
    \fbox{$\Rightarrow$}
    Suppose $\varphi$ is satisfiable. Hence there exists a small filtrated model $\M = \ldiamondarg{S, \{R_i\}_{i\in \Ag}, V, Exp}$ such that $\M,s\vDash\varphi$ for some $s\in S$ and $|S|\leq 2^{FL(\varphi)}$ (by Theorem \ref{theorem:filtration},). Note that, in the context of this proof, $\M|_w = \ldiamondarg{S^w, \{R^w_i\}_{i\in \Ag}, V^w, Exp^w}$ for any $w\in\Sigma^\star$.


    We now create a transition system $\B_\varphi = \ldiamondarg{\Bb,\delta}$ and prove it to be the finite transition system for $\varphi$.

    \begin{itemize}
        \item For all $w\in\Sigma^\star$, we set $B^w := \ldiamondarg{S^w, \{R^w_i\}_{i\in \Ag}, L^w}$ such that for any $s\in S^w$, $L^w(s) = \{\psi\in FL(\varphi)\mid \M|_w,s\vDash\psi\}$.
        
        \item $\Bb = \{B^w \mid  w\in\Sigma^\star\}$
        \item $$\delta(B, a) = 
        \begin{cases}
            \set{B^{wa}} \text{, $\exists~ w$ such that $B = B^w$ and $\M|_{wa}$                                       exists} \\
            \emptyset \text{ otherwise}
        \end{cases}$$
    \end{itemize}

    Now we prove that the transition structure satisfies every point in Definition \ref{defi:bubbletrans}.
    \begin{enumerate}


            \item \textbf{Each $B^w\in\Bb$ is a bubble.} $B^w = \{S^w,\{R^w_i\}_{i\in \Ag}, L^w\}$. 
            \begin{itemize}
                \item Consider any $s\in S^w$.We show proof when $K_i\psi\in L^w(s)$ and $\ldiamondarg{\pi^\star}\psi\in L^w(s)$. Since $\M|_w,s\vDash K_i\psi$ and the indistinguishability relations are also reflexive, hence $\M|_w,s\vDash\psi$, hence $\psi\in L^w(s)$. Now consider $\M|_w,s\vDash\ldiamondarg{\pi^\star}\psi$, hence by truth condition $\M|_w,s\vDash\ldiamondarg{\pi}\ldiamondarg{\pi^\star}\psi$. Hence $\ldiamondarg{\pi}\ldiamondarg{\pi^\star}\psi\in L(s)$. With similar deductions it can be proved for each $s\in S^w$, $L^w(s)$ is a Hintikka set.
    
                \item Since it is proved each $s\in S^w$ is such that $L^w(s)$ is a Hintikka set, we now prove $B^w = \ldiamondarg{S^w, \{R^w_i\}_{i\in \Ag}, L^w}$ is a bubble. Consider a formula $\hat{K_i}\psi$ such that $\M|_w,s\vDash\hat{K_i}\psi$. Hence there is another $s'$ such that $s R^w_i s'$ and $\M|_w,s'\vDash\psi$. Hence there a $s'\in S^w$ such that $s R^w_i s'$ and $\psi\in L^w(s')$. The knowledge condition can also be proved similarly using the fact that $R_i$ is an equivalence relation.
            \end{itemize}
        
        \item Since $\M$ is such that $\M,s\vDash\varphi$, we have our second condition of definition \ref{defi:bubbletrans}.


        \item Consider $B^w$ and by construction $\delta(B^w, a) = \{B^{wa}\}$ if $\M|_{wa}$ exists (implying $\M|_w$ exists as well). Since valuation in a state of the model remains consistent in updates, condition 1 of definition \ref{defi:ObsSucc} is satisfied. $\M|_w,s\vDash\ldiamondarg{a}\psi$ iff $\M|_{wa},s\vDash\psi$, hence satisfying condition 2 (similar argument for condition 3). Since distinguishability relation disappears whenever a state disappears and a non-surviving state can never reappear in further updates, condition 4 is satisfied.


        \item Let $\ldiamondarg{\pi}\psi\in L^w(s)$. Since $\M|_w,s\vDash\ldiamondarg{\pi}\psi$ implies there exists a $w'\in\LL(\pi)$ such that $s$ survives in $\M|_{ww'}$ and $\M|_{ww'},s\vDash\psi$, hence there exists a series of $\delta$ transitions labelled by $w'$ after which $\psi\in L^{ww'}(s)$.
    \end{enumerate}
\end{proof}

\section{Proofs from Section 5}
\dpdltranscorrect* 
\begin{proof}
\newcommand{\statefor}[1]{s_{#1}}
    ($\implies$)
     Given a satisfiable formula $\phi$, there is a $\FBTS$ $\B_\phi = \ldiaarg{\Bb,\delta}$ of $\varphi$ as per \Cref{theorem:transmodelsound}. Now we build a DPDL model $\modelDPDL = \ldiaarg{W, \{\rightarrow_a\}_{a\in\Sigma}, V}$ of $tr(\phi)$ out of $\B_\phi$, such that $\modelDPDL,s\vDash tr(\phi)$ for some $s\in W$.
     \begin{itemize}
        \item \textbf{The States out of bubble.} 
        For all bubbles $B\in\Bb$, we introduce a state $\statefor B \in W$. The states $\statefor B$ are pair-wise distinct. We set $W = \{\statefor B \mid B\in\Bb\}$.

        \item \textbf{The transitions from Observation Successor.} 
      We define the deterministic transition relation $\rightarrow_a$ so that it mimics $\delta$. That is, $\statefor B \rightarrow_a \statefor{\delta(B,a)}$ if $\delta(B,a) \neq \nosuccessor$, 
        while there is no $s'$ such that $\statefor B \rightarrow_a s'$ when $\delta(B,a) = \nosuccessor$.
        
        \item \textbf{Valuations encode Kripke Structure. }
        Here we define propositions for each state and the formula the state satisfies as propositions that look like $@\labelworld.\psi$ such that $\labelworld$ is a state in a bubble and $\psi$ is a formula it has. Formally, for a bubble $B = \ldiaarg{S, \{R_i\}_{i\in\Ag}, L}$, 
        we define $V(s_B) = \{@\labelworld.\psi\mid \labelworld\in S, \psi\in L(\labelworld)\}\union\{surv(\labelworld)\mid \labelworld\in S\}\union\{R_i(\labelworld,\labelworld^\prime)\mid (\labelworld,\labelworld^\prime)\in R^\star_i\}$ where $R^\star_i$ is the relation in~$B^\star$.
     \end{itemize}
     By definition of $\FBTS$, there is a bubble $B^\star$ and a state $1$ in it which has $\phi$ labeled in it. Therefore by construction, $\modelDPDL,s_{B^\star}\vDash @1.\phi\wedge surv(1)$. Now we prove the formulas of $\cS_\phi$ number by number:
     \begin{enumerate}
        \item We do a case by case analysis on the structure of $\psi$ to prove $\modelDPDL,\statefor B \vDash\translationsem(\phi)$ at any label of any bubble $B$. We show the complicated case of $\phi = \hat{K}_i\psi$ and $\phi = \ldiaarg{\pi}\psi$. For the first case $\translationsem(\hat{K}_i\psi) = \bigwedge_{l \in \setlabels\phi} (@\labelworld.\hat{K}_i\psi\leftrightarrow
        (\bigvee_{l^\prime \in \setlabels\phi}(\proprel i \labelworld {\labelworld^\prime}\wedge surv(\labelworld^\prime)\wedge @\labelworld^\prime.\psi)))$. If $\hat{K}_i\psi\in L(\labelworld)$ of bubble $B$, then by construction $\modelDPDL,\statefor B\vDash @\labelworld.\hat{K}_i\psi$. Because of condition 3(a) of $\FBTS$ (definition~\ref{def:epibubble}) and by construction, $\modelDPDL,\statefor B\vDash \proprel i \labelworld {\labelworld^\prime}\wedge surv(\labelworld^\prime)\wedge @\labelworld^\prime.\psi$. Suppose $\hat{K}_i\psi\notin L(\labelworld)$, hence similarly due to 3(a) and 3(b) of definition~\ref{def:epibubble}, $\modelDPDL,\statefor B \vDash\neg @\labelworld.\hat{K}_i\psi$ and also $\modelDPDL,\statefor B \nvDash \proprel i \labelworld {\labelworld^\prime}\wedge surv(\labelworld^\prime)\wedge @\labelworld^\prime.\psi$ for any $\labelworld^\prime$ in $B$. 

        Now consider $\phi = \ldiaarg{\pi}\psi$. Suppose $\ldiaarg{\pi}\psi\in L(\labelworld)$. Hence by construction $\modelDPDL,\statefor B \vDash @\labelworld.\ldiaarg{\pi}\psi$. Also, due to condition (4) of $\FBTS$ definition~\ref{defi:bubbletrans} and by construction $\modelDPDL,s(B_k)\vDash @\labelworld.\psi\wedge surv(\labelworld)$ such that $\statefor B \rightarrow_{w}^\star s(B_k)$ where $w\in\LL(\pi)$. Therefore $\modelDPDL,\statefor B \vDash\ldiaarg{\pi}(@\labelworld.\psi\wedge surv(\labelworld))$. Similarly $\modelDPDL,\statefor B \nvDash\ldiaarg{\pi}(@\labelworld.\psi\wedge surv(\labelworld))\wedge @\labelworld.\ldiaarg{\pi}\psi$ when $\ldiaarg{\pi}\psi\notin L(\labelworld)$.

        \item This condition is satisfied because of the property (2) of observation successor (\Cref{defi:ObsSucc}).
        
        \item Formula 3 is true in $M, s_{B^\star}$ ..... due to perfect recall property (5) of observation successor (\Cref{defi:ObsSucc}).

        \item This formula is satisfied due to the fact that $R_i$ relations are equivalence.
        
        \item This formula is true from construction. A proposition of form $@\labelworld.\psi$ can only belong in a DPDL state when there was a $\labelworld$ which has $\psi$ labeled in it. This implies $\labelworld$ survived in that bubble $B$.
        
        \item This is due to condition (1) of \Cref{defi:ObsSucc}.
     \end{enumerate}

    ($\impliedby$)
    Assume that there is a DPDL model $\modelDPDL,s\vDash tr(\varphi)$.
    We define an $\FBTS$ of $\phi$ from $\modelDPDL = \ldiaarg{W, \{\xrightarrow{a}\}_{a\in\Sigma}, V}$ satisfying $tr(\varphi)$.

    We take each $w\in W$ and build a bubble $B(w) = \ldiaarg{S, \{R_i\}_{i\in\Ag, L}}$ out of it:
    \begin{enumerate}
        \item $S = \{\labelworld\mid \mbox{there is some }\psi:@\labelworld.\psi, surv(\labelworld)\in w\}$
        \item $R_i = \{(\labelworld,\labelworld^\prime)\mid surv(\labelworld), surv(\labelworld^\prime), \proprel i \labelworld {\labelworld^\prime}\in w\}$
        \item $L(\labelworld) = \{\psi\mid @\labelworld.\psi\in w\}$ 
    \end{enumerate}
    \noindent\paragraph{Is $B(w)$ a bubble?} The condition 1 and 2 satisfies by filtration and definition. For $3(a)$, take any $\labelworld\in S$ such that some $\hat{K}_i\psi\in L(\labelworld)$. Note the translation $\translationsem(\hat{K}_i\psi) = \bigwedge_{l \in \setlabels\phi} (@\labelworld.\hat{K}_i\psi\leftrightarrow
    (\bigvee_{l^\prime \in \setlabels\phi}(\proprel i \labelworld {\labelworld^\prime}\wedge surv(\labelworld^\prime)\wedge @\labelworld^\prime.\psi)))$. Hence, by the second part after $\leftrightarrow$ and by the construction of bubble, there is another $\labelworld^\prime$ such that $\proprel i \labelworld {\labelworld^\prime}$ such that $\psi\in L(\labelworld^\prime)$. Similarly, $3(b)$ can also be verified.

    Let $\B(\modelDPDL) = \ldiaarg{\Bb(W),\delta}$ be the $\FBTS$ where $\Bb(W)$ is the set of all bubbles out of every $w\in W$, and $\delta = \{B(s)\xrightarrow{a} B(t)\mid s\xrightarrow{a} t \text{in $M$}\}$.

    \noindent\paragraph{Is $B(t)$ an $a$-observation successor of $B(s)$?} The point 8 of the structural formula $\cS_\phi$ ensures that once a label $\labelworld$ does not survive ($\neg surv(\labelworld)$), it can never grow back in future ($\ldiaarg{a}\neg surv(\labelworld)$). Hence, unique labels will remain same or will decrease in number. The second condition of observation successor is enforced by the second point which says the proposition valuation of surviving labels will stay contant as observation occurs. 

    Suppose a formula $\ldiaarg{\pi}\psi\in FL(\phi)$ such that $\pi\setminus a\neq\emptyset$ and $\ldiaarg{\pi}\psi\in L_s(\labelworld)$, where $L_s$ is the labelling function of $B(s)$ and $\labelworld$ is some label in $B(s)$. As per how $FL(\phi)$ breaks formulas down, and since $\pi\regdiv a\neq\emptyset$, therefore $\ldiaarg{a}\ldiaarg{\pi\regdiv a}\psi\in FL(\phi)$. Therefore by construction $@\labelworld.\ldiaarg{a}\ldiaarg{\pi\regdiv a}\psi\in s$. Recall $\translationsem(\ldiaarg{a}\ldiaarg{\pi\regdiv a}\psi) = \\
    \bigwedge_{l \in \setlabels\phi} (@\labelworld.\ldiaarg{a}\ldiamondarg{\pi\regdiv a}\psi
    \leftrightarrow\ldiamondarg{a}(@\labelworld.\ldiaarg{\pi\setminus a}\psi\wedge surv(\labelworld)))$. Therefore $\ldiamondarg{a}(@\labelworld.\ldiaarg{\pi\setminus a}\psi\wedge surv(\labelworld))\in s$ which means, by construction of $B(t)$, since $@\labelworld.\ldiaarg{\pi\setminus a}\psi, surv(\labelworld)\in t$, $\labelworld$ survives in $B(t)$ and $\ldiaarg{\pi\setminus a}\psi\in L_t(\labelworld)$, where $L_t$ is the labelling function for $B(t)$. Similarly, the converse, that is assuming $\ldiaarg{\pi\regdiv a}\psi\in L_t(\labelworld)$ and $\labelworld$ is in $B(t)$, it can be proven $\ldiaarg{\pi}\psi\in L_s(\labelworld)$ which altogether proves the satisfiability of the condition (3) of observation successor.

    In a similar way, condition (4), which is the condition of box formulas, can also be proven.

    Condition (5) is perfect recall which is enforced by the third point in the definition of $\cS_\phi$ formula.

    \noindent\paragraph{Condition (4) of $\FBTS$. } Consider a formula $\ldiaarg{\pi}\psi\in L_s(\labelworld)$, where $L_s$ is the label function of some $B(s)$ which is some bubble corresponding to $s$ in $\modelDPDL$. Recall the translation $\translationsem(\ldiamondarg{\pi}\psi) = \bigwedge_{l \in \setlabels\phi} (@\labelworld.\ldiamondarg{\pi}\psi
    \leftrightarrow\ldiamondarg{\pi}(@\labelworld.\psi\wedge surv(\labelworld)))$. Since $\ldiaarg{\pi}\psi\in L_s(\labelworld)$, hence $@\labelworld.\ldiaarg{\pi}\psi\in s$. And as per $\translationsem(\ldiamondarg{\pi}\psi)$, $\modelDPDL,s\vDash\ldiamondarg{\pi}(@\labelworld.\psi\wedge surv(\labelworld))$. Hence there is some $w\in\LL(\pi)$, such that $s\xrightarrow{w}^* t$ and $\modelDPDL,t\vDash @\labelworld.\psi\wedge surv(\labelworld)$. The last point formula of $\cS_\phi$ ensures once a $surv(\labelworld)$ turns false, it can never be turned true. Hence, in $s$, the label $\labelworld$ survives and keeps on surviving till $t$ on the path $s\xrightarrow{w}^* t$.

    
\end{proof}

\section{Proofs from Section~\ref{sec:lowerbound}}

\lowerboundtrpolytime*

\begin{proof}
    Give $x$, we compute $e(|x|)$ and $n$ is the number of bits used to represent $e(|x|)$. The number $n$ is polynomial in $|x|$. Formulas (1-\ref{equation:lowerboundlastequation}) are then computable in poly-time in $|x|$.
\end{proof}

\lowerboundtrcorrect*

\begin{proof}
    The instance $x$ is positive iff there is an accepting computation tree of the machine $M$ on $x$. We now prove that  there is an accepting computation tree of the machine $M$  on $x$ iff $tr(x)$ is POL-satisfiable.

    Figure~\ref{fig:hardnessbigpicture} sums up the main idea of a correspondence between a computation tree of $M$ on $x$ and its pointed corresponding POL-model $\modelM, s$ that satisfies $tr(x)$.

The pointed corresponding POL-model $\modelM, s$ has a specific form, we call it a pointed \emph{POL computation tree model} meaning that it encodes a computation tree by satisfying the following constraints (we use the same numbering than the formulas 1-\ref{equation:lowerboundlastequation} in the reduction):
\begin{itemize}
    \item (1)
    The model contains a full binary tree for modalities $\ldia$ and $\lbox$ of depth $3n$. It branches over the propositions $p_\ell$. At the root, we branch over the two values of $p_1$. At the children of the root, we branch over the two values of $p_2$, etc. At the parent of the leaves, we branch over the two values of $p_{3n}$.
    \item (2-3) Whatever has been observed so far, values of propositions $p_m$ are observable at any node of the tree. If $p_m$ is true, then the observation $p_m$ is observable. If $p_m$ is false then the observation $\bar{p_m}$ is observable.
    \item (4) After observing any sequence of $a$ and $b$, at each leaf, there is exactly one $\alpha$ such that $1{:}\alpha$ is observable there, exactly one $\alpha$ such that $2{:}\alpha$ is observable there, and exactly one $\alpha$ such that $3{:}\alpha$ is observable there.
    \item (5-6) The observability of some $i{:}\alpha$ remains unchanged when observing a sequence of $p_\ell$ or $\bar{p_\ell}$.
    \item (7) For all position $pos_1$, all leaves for the same position $pos_1$ for the first configuration have the $1{:}\alpha$ that is observable. Same for the second and third configuration.
    \item (8) The three superposed configurations are the same.
    \item (9) Observability of $a$ and $b$ is uniform in the complete binary tree, (non-)observability of $a$ is the same for all nodes of the tree.
    \item (10-11) $1{:}\leftmostrightmostsymbol$ is observable when the position in the first configuration is extremal (0 or $e(|x|) - 1$).
    \item (12-13) When no $a$ and $b$ have been observed (i.e. in the complete binary tree corresponding to the initial configuration), the first configuration written at the leaves is the initial configuration.
    \item (14) If, after observing any sequence of $a$ and $b$, we reach a tree that corresponding to a non-final configuration, then the next tree after observing $a$ contains the next configuration via an $a$-transition. Same for $b$.
    \item (15) If, after observing any sequence of $a$ and $b$, we reach a final state, then $a$ and $b$ is not observable anymore.
    \item (16-17) The observability of $\exists$ is swapped after observing $a$ or $b$. 
    \item (18-19) After observing a sequence of $a$ and $b$, if the tree represents an accepting (resp. rejecting) configuration then $win$ is observable (resp. not observable).
    \item (20-21) After observing a sequence of $a$ and $b$, if the three represents a non-terminal configuration then the winning condition is existential or universal depending on whether $\exists$ is observable or not.
    \item (22) Both $win$ and $\exists$ are observable.
\end{itemize}

    We suppose that each non-terminal configuration in the computation tree has a left ($a$-) and a right ($b$-) child.
Recall that $n$ is the number bits needed to encode a position on the tape, that is a number in $\set{0, \dots, e(|x|)}$ where $e$ is an exponential function.

    \fbox{$\Rightarrow$} Consider an accepting computation tree of $M$ on~$x$. We construct a pointed POL computation tree model $\modelM, s$ that satisfies $tr(x)$ as follows.
    
    The expectation function is defined as follows: at a given leaf $\leaf$, 
    $Exp(\leaf)$ contains exactly the word $d_1\dots d_k i{:}\alpha$ such that $d_1, \dots d_k \in \set{a, b}$ if following $d_1, \dots, d_k$ in the computation tree from the root leads to a configuration in which $\alpha$ is written in the cell at index represented by $p_{(i-1)\times n + 1}, \dots, p_{(i-1)\times n + n}$. 
    Note that $Exp(\leaf)$ is finite, thus regular. 
    By construction, the model at the root satisfies $tr(x)$.

    \fbox{$\Leftarrow$} Conversely if $tr(x)$ is satisfiable, by definition of $tr(x)$ there a pointed POL computation tree model $\modelM, s$ of $tr(x)$. We extract an accepting computation tree of $M$ on $x$ as follows.

    The configuration obtained from the root by taking the directions $d_1, \dots, d_k \in \set{a, b}$ is defined as follows. Let us explain how to get the symbol written at a given position on the cell. Consider a position whose binary representation is $b_1\dots b_n$. We first look into $M_{\mid d_1, \dots, d_k}$. We then follow the $\ldia$-path in the binary tree in $M_{\mid d_1, \dots, d_k}$: going in the $\ldia$-successor where $p_1$ holds if $b_1$ is true, going in the $\ldia$-successor where $p_1$ does not hold if $b_1$ is true, 
    $\dots$ going in the $\ldia$-successor where $p_n$ holds if $b_n$ is true, going in the $\ldia$-successor where $p_n$ does not hold if $b_n$ is true. Then take any $\ldia$-successor until reaching a leaf. We then consider the unique symbol $\alpha$ such that $1{:}\alpha$ is observable in that leaf: $\alpha$ is the symbol written in that position.

    By construction, the constructed computation tree is accepting. Indeed, by induction on the depth of the subtree, each subtree by taking directions $d_1, \dots, d_k \in \set{a, b}$ has a winning configuration iff $\modelM, s \models \ldiaarg{d_1\dots d_k} \propwin$.
\end{proof}

\end{document}